\documentclass[10pt,journal,compsoc]{IEEEtran}

\usepackage{amssymb}
\usepackage{subcaption}
\usepackage[noend]{algpseudocode}
\usepackage{multirow}
\usepackage{bbm}
\usepackage{enumitem}
\usepackage{amsmath}
\usepackage{bbm}
\usepackage{color}
\usepackage[pdftex]{graphicx}
\usepackage{booktabs}
\usepackage{siunitx}
\usepackage{textcomp}
\usepackage{mdwlist}
\usepackage{xcolor}

\ifCLASSOPTIONcompsoc
  \usepackage[nocompress]{cite}
\else
  \usepackage{cite}
\fi

\usepackage{times}  
\usepackage{helvet} 
\usepackage{courier}  
\usepackage[hyphens]{url}  
\usepackage{algorithm}

\newcommand{\indep}{\rotatebox[origin=c]{90}{$\models$}}

\usepackage{amsmath}
\begin{document}
%
\title{Towards Causality-Aware Inferring: A Sequential Discriminative Approach for Medical Diagnosis}

\author{Junfan Lin,
        Keze Wang,
        Ziliang Chen,
        Xiaodan Liang,
        and Liang Lin
\thanks{J. Lin, K. Wang, and L. Lin are with the School of Computer Science and Engineering, Sun Yat-sen University, Guangzhou, China, 510006. And X. Liang was with the School of Intelligent Systems Engineering, Sun Yat-sen University, Guangzhou, China, 510006. J. Lin, K. Wang, X. Liang, and L. Lin are also with the Engineering Research Center for Advanced Computing Engineering Software of the Ministry of Education, China. Z. Chen was with the Guangdong Institute of Smart Education, Jinan University. Email: (Email: linjf8@mail2.sysu.edu.cn, kezewang@gmail.com, zlchilam@163.com, xdliang328@gmail.com, linliang@ieee.org)
}
\thanks{Corresponding author: Keze Wang}
\thanks{Manuscript received June 30, 2022; revised February 16, 2023; revised May 26, 2023.}}%
%
%

\markboth{IEEE Transactions on Pattern Analysis and Machine Intelligence}%
{Shell \MakeLowercase{\textit{et al.}}: Bare Demo of IEEEtran.cls for Computer Society Journals}
%



\IEEEtitleabstractindextext{%
\begin{abstract}
Medical diagnosis assistant (MDA) aims to build an interactive diagnostic agent to sequentially inquire about symptoms for discriminating diseases. However, since the dialogue records for building a patient simulator are collected passively, the collected records might be deteriorated by some task-unrelated biases, such as the preference of the collectors. These biases might hinder the diagnostic agent to capture transportable knowledge from the simulator. 
This work identifies and resolves two representative non-causal biases, i.e., (i) default-answer bias and (ii) distributional inquiry bias.
Specifically, Bias (i) originates from the patient simulator which tries to answer the unrecorded inquiries with some biased default answers. {To eliminate this bias and improve upon a well-known causal inference technique, i.e., propensity score matching, we propose a novel propensity latent matching in building a patient simulator to effectively answer unrecorded inquiries;} Bias (ii) inherently comes along with the passively collected data that the agent might learn by remembering what to inquire within the training data while not able to generalize to the out-of-distribution cases. To this end, we propose a progressive assurance agent, which includes the dual processes accounting for symptom inquiry and disease diagnosis respectively. The diagnosis process pictures the patient mentally and probabilistically by intervention to eliminate the effect of the inquiry behavior. And the inquiry process is driven by the diagnosis process to inquire about symptoms to enhance the diagnostic confidence which alters as the patient distribution changes. In this cooperative manner, our proposed agent can improve upon the out-of-distribution generalization significantly. Extensive experiments demonstrate that our framework achieves new state-of-the-art performance and possesses the advantage of transportability. The source code is available at \url{https://github.com/junfanlin/CAMAD}.
\end{abstract}

}

\maketitle

\IEEEdisplaynontitleabstractindextext

%
\IEEEpeerreviewmaketitle

\IEEEraisesectionheading{\section{Introduction}\label{sec:introduction}}
A medical diagnosis assistant~(MDA) aims to learn an active agent from passively collected data, which can be used to collect symptom information and make preliminary diagnoses. Specifically, the MDA agent sequentially inquires about symptom information from the user/patient, and proactively terminates the interaction by informing the discriminant diagnosis. Such sequential discrimination has been formulated as a Markov decision process and is resolved by reinforcement learning~(RL)~\cite{mnih2015human,sutton1999policy,xu2019end,wei2018task}. However, we found that the previous RL-based MDA methods~\cite{xu2019end,wei2018task} have omitted several causal biases in building the patient simulator and designing the diagnostic agent. Learning MDA agents without mitigating these biases would hamper the agent to discover the causal and transportable skills of interest behind the data. According to the sources of these biases, we denote them as \textit{default-answer bias} and \textit{distributional inquiry bias}, respectively. 

\begin{figure}[t]
	\centering
	\includegraphics[clip=true, trim=0pt 0pt 0pt 10pt, width=0.9\linewidth]{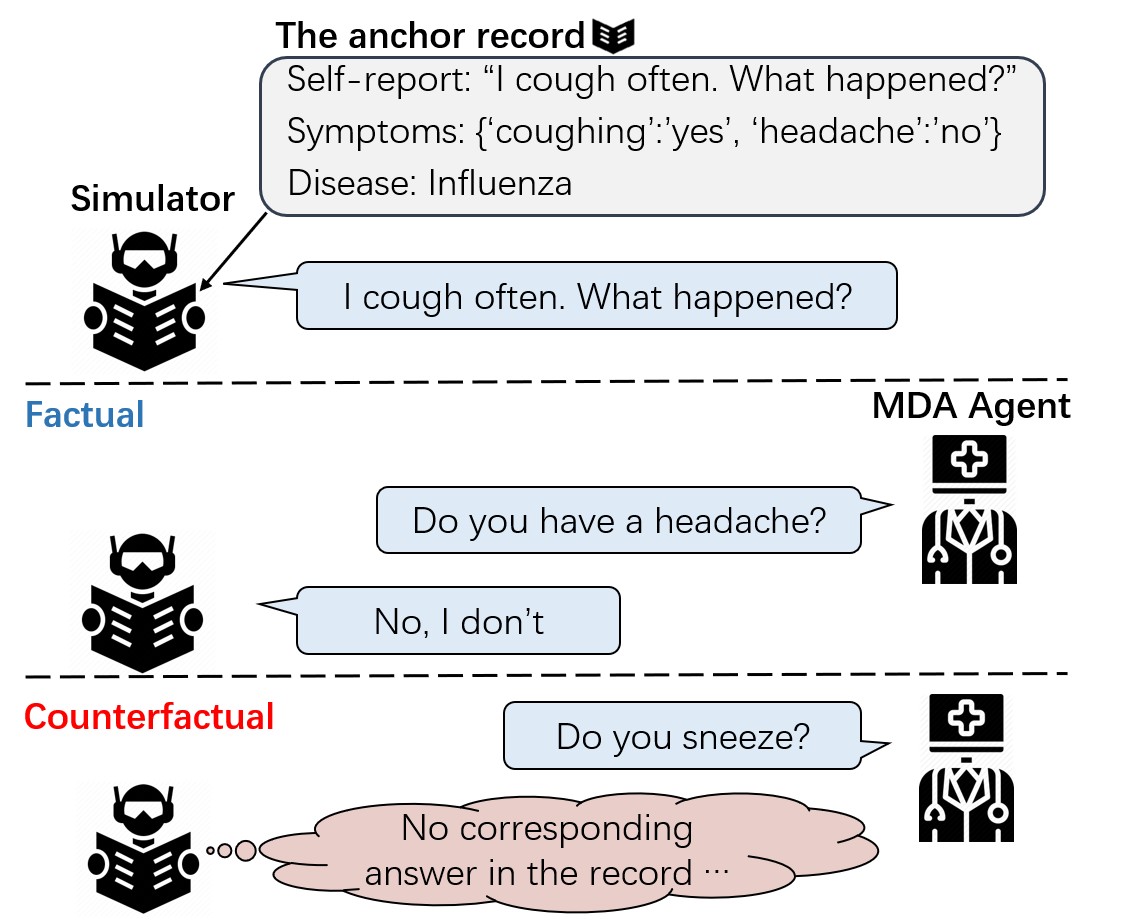}
	\vspace{-3mm}
	\caption{The patient simulator chooses a record as the anchor record and gives its self-report to start a diagnosis process. Then, the simulator answers the \emph{factual symptom inquiries} which are already observed in the anchor record but would fail to answer the \emph{counterfactual symptom inquiries} about the unobserved symptoms.}
	\vspace{-15pt}
	\label{fig:cfd}
\end{figure}

\begin{figure}[t]
	\centering
	\includegraphics[clip=true, trim=0pt 200pt 470pt 35pt, width=0.95\linewidth]{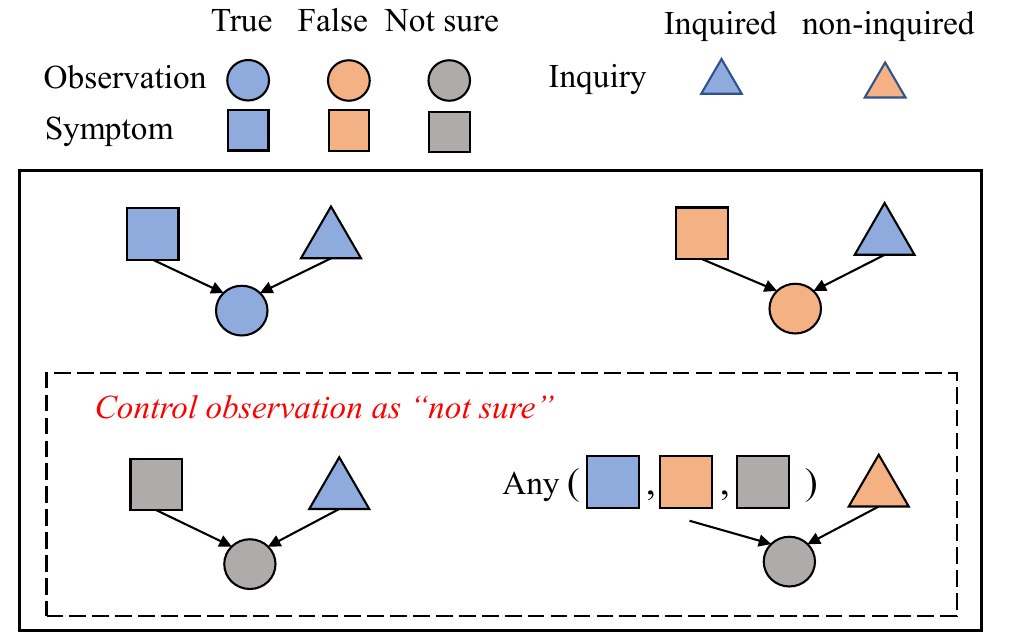}
	\vspace{-2mm}
	\caption{The causal diagrams among the symptom, observation, and inquiry. Causally~(boxed by solid line), the status of the symptom is independent of the status of the inquiry. When the observation is controlled as ``not sure"~(boxed by dashed line), the status of the symptom is biased and becomes dependent on the status of the inquiry.}
	\label{fig:collider}
	\vspace{-10pt}
\end{figure}
 
\textbf{Default-answer bias.} In prior methods~\cite{xu2019end,wei2018task}, the patient simulator will choose a dialogue diagnosis record from the collected dataset as the anchor record (as exemplified in the top of Fig.~\ref{fig:cfd}), and then answer the inquiry from the MDA agent by looking the answer up in the anchor record. However, as the passive record only reflects the \textit{factual} side of the world, the simulator might fail to answer the unrecorded inquiries~(i.e., the \textit{counterfactual} aspects) from the agent, illustrated in Fig.~\ref{fig:cfd}. To deal with the counterfactual inquiries, prior works~\cite{wei2018task,xu2019end} make the simulator render `not sure' responses as the default answers. Unfortunately, the default-answer strategy will bring about the collider/selection bias~\cite{cole2010illustrating} among the symptom, inquiry, and answer, named default-answer bias in our paper. As depicted in Fig.~\ref{fig:collider}, by convention, the answer/observation is the result/collider of the \textit{inquiry} from the agent and \textit{symptom} of the patient. As shown in the dashed box of Fig.~\ref{fig:collider}, controlling the observation to be `not sure' will result in a biased association that the status of the symptom is dependent on the status of the inquiry. This bias hampers the agent from learning the causal relationship between symptoms and diseases. Worse, due to the sparsity of the symptom information in MDA, the simulator will be frequently inquired about unrecorded symptoms during the training phase, amplifying the collider bias's negative impact. Training and evaluating under the simulator with severe collider bias cannot fully reflect the advantages of the diagnostic agent.

To reduce the default-answer bias, the simulator should infer the symptom ignoring the inquiry and render the answer accordingly rather than providing the default answer to the counterfactual inquiry. 
In the field of causal inference, one of the most popular counterfactual inference methods to provide such `ignorability'~\cite{rubin1974estimating,neyman1923applications} is \textit{Propensity Score Matching} (PSM)~\cite{dehejia2002propensity,austin2011introduction,rosenbaum1983central} from the potential outcome framework~(PO). In our context, PSM is also suitable to address the sparsity problem by abstracting the sparse covariates to low-dimensional propensity scores via a propensity score estimator. However, during the interactive learning process of the MAD agent, counterfactual inquiries are frequently raised. The counterfactual inquiry might be out-of-distribution to the propensity score estimator, resulting in unreliable propensity score estimation. 
To this end, our paper proposes a novel \textit{Propensity Latent Matching}~(PLM), which conducts matching in the latent space. Unlike PSM which predicts the propensity score after seeing the counterfactual inquiry, our PLM models the latent features before the counterfactual inquiry gets involved in estimating the propensity, which makes the matching more reliable.

\begin{figure}[t]
	\centering
    \includegraphics[clip=true, trim= 0pt 0pt 0pt 10pt, width=\linewidth]{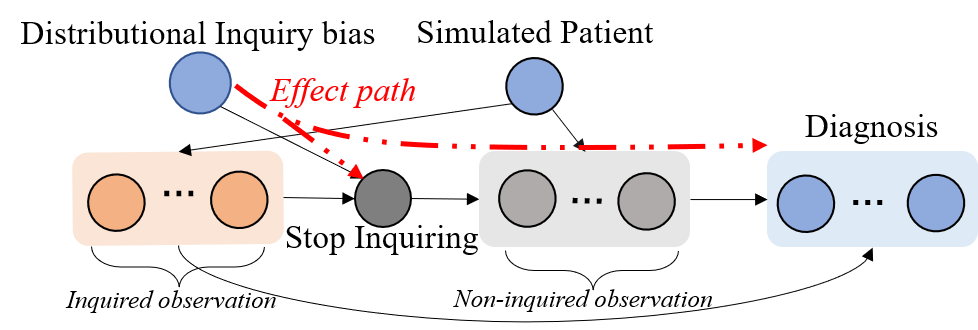}
\vspace{-5mm}
	\caption{The causal diagrams for the diagnostic agent. 1) The distributional inquiry bias misleads the agent to stop inquiry with insufficient symptom observation; 2) Furthermore, insufficient observations might cause the diagnoser to overfit the training distribution.}
		\label{fig:p2a_total}
	\vspace{-6mm}
\end{figure}

\textbf{Distributional inquiry bias.} Another representative bias that comes along with passive data is the distributional bias, which is also one of the key obstacles for training the agent towards “learning how” rather than “remembering what”. Due to the difficulty and high cost of collecting MDA data, the diagnostic agent suffers from a limited amount of training data and might be fragile in out-of-distributional cases. For example, if inquiring about the symptom `coughing' is enough for distinguishing between two diseases among the training samples, the agent tends to learn a `shortcut' that only inquires about `coughing' which also hampers the learning of discrimination. We name such kind of bias as the \textit{distributional inquiry bias}. As shown in Fig.~\ref{fig:p2a_total}, the distributional inquiry bias not only affects the inquiry process (i.e., the effect on `stop inquiring') but also affects the learning of diagnosis mediated by the `non-inquired observations'. In this manner, at the deployment phase where more symptom inquiries are needed, the agent might be unable to collect enough information and make reliable diagnoses.

\begin{figure*}[t]
	\centering
	\includegraphics[clip=true, trim=0pt 365pt 0pt 0pt, width=\linewidth]{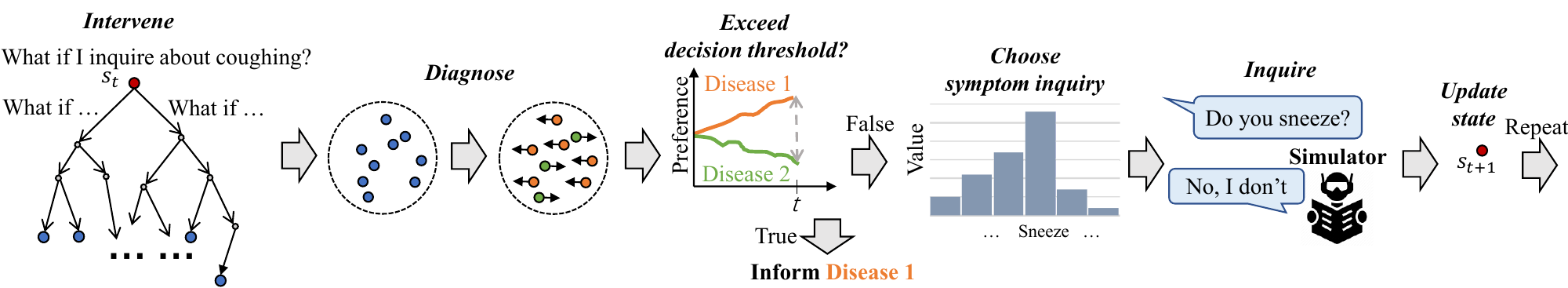}
	\vspace{-5mm}
	\caption{An example sketch of our P2A with binary disease alternatives. P2A conducts intervention to picture the complete patient and diagnose all possible imagining patients. After, it estimates the timely confidence of the two disease candidates. Once the confidence gap between two alternatives exceeds the decision threshold, the winner with higher confidence will be executed. Otherwise, P2A continues inquiring about symptoms to collect more evidence.}
	\label{fig:p2a_dif}
	\vspace{-5mm}
\end{figure*}

To mitigate these issues, we propose a novel MDA agent, progressive assurance agent~(P2A), which leverages the confidence of diagnosis to modulate the inquiry process and conducts the intervention to improve the robustness of the diagnosis. In specific, P2A is a dual-process agent, consisting of two separate yet cooperative branches, i.e., the diagnosis branch and the inquiry branch. The diagnosis branch takes advantage of the intervention/do-calculus~\cite{Pearl2012TheDr}, i.e., do(Causes of non-inquired observations = simulated patient), to \textit{cut off the effect path from distributional inquiry bias to the diagnosis}. And the inquiry branch is modulated by the diagnosis branch to collect evidence to enhance the diagnostic confidence until the decision threshold is met. The diagnostic confidence can sensitively reflect the distribution changes and thus \textit{adapt the inquiry branch to out-of-distribution scenarios}. 
Taking a binary-alternatives case as an example (Fig.~\ref{fig:p2a_dif}), the diagnosis branch first reasons and plans according to historical observations (namely, $s_t$ in the figure) by intervening in the unknown aspects, i.e., the non-inquired observations, of the patient. By intervention, the diagnostic decision in our P2A is made according to the more comprehensive and imaginary patient information instead of the observed information limited to the inquiry behavior. Then, the imagined patients are diagnosed into different disease clusters (in either orange or green). Once the population of one cluster is significantly overwhelming the other by a decision threshold~\cite{ratcliff2016diffusion}, the agent will stop inquiring and the corresponding disease of that cluster will be informed.  Particularly, the competitive behavior between different clusters and the concept of the decision threshold of our P2A are all derived from the diffusion model in neuroscience~\cite{ratcliff2016diffusion}. Not limited to the binary case in neuroscience, our P2A has extended the diffusion model to a multi-alternatives setting from a computational perspective.

Overall, the main contributions of our paper are tri-fold: i) We identify the default-answer bias that existed in the previous patient simulator and {propose a propensity-based patient simulator~(PBPS) using the novel propensity latent matching}; ii) we also identify the distributional inquiry bias and propose a novel MDA agent, progressive assurance agent, to eliminate the distributional bias via intervention; iii) we develop a neuro-inspired `decision threshold' mechanism for addressing the sequential discrimination problems, which seamlessly connects the inquiry and discrimination to make reliable and interpretable decisions. Considering the addressed causality issues, we name our framework, consisting of PBPS and P2A, as Causality-Aware MDA (CA-MDA). Experimental results demonstrate that: i) our PBPS is superior in answering counterfactual symptom inquiry and generating more informative answers; ii) our P2A advances in capturing the symptom-disease relationship and generalizing it to out-of-distribution cases, and also possesses the advantage of sample-efficient and robustness. 

The remainder of this paper is organized as follows. Section 2 comprehensively reviews the related works. Section 3 presents the background of CA-MDA and the limitation of RL-based formulation. Section 4  elaborates on our CA-MAD and Section 5 shows the experimental results and human evaluation of two public MDA benchmarks with comprehensive evaluation protocols. Section 6 concludes this paper.

\section{Related Work}

In the field of causal inference, there are many different methods proposed in recent years, which can be approximately categorized into two branches according to whether the causal relationship is modeled explicitly or implicitly, namely, the structural causal model (SCM) framework~\cite{pearl2009causal} and the potential outcome framework~\cite{rubin1974estimating}. The methods based on the SCM usually incorporate a causal diagram~(CD) and structural causal equations~\cite{bollen2013eight}. Then, the collected data are injected into the model to infer the causal effect. Recently, many approaches based on the CD have been proposed in computer vision~\cite{chen2019counterfactual,kaihua2020unbiased,qi2020two,wang2020visual,abbasnejad2020counterfactual}, reinforcement learning~\cite{yu2021reinforcement,dasgupta2018causal,oberst2019counterfactual,foerster2018counterfactual}, etc. Most of these works are based on the general CD for the population. In MDA, patients with the same diseases may have different symptoms due to individual differences. As for modeling the individual-level relation, the model-free causal inference framework~(i.e., the potential outcome framework~\cite{rubin1974estimating} (PO)) is widely adopted. PO infers the causal effect from passive observational data without an explicit causal model and it's also capable of handling individual causal effects. 

In MDA, the collected dialogue records are passively observed, that's, the existence of the unobserved symptom is \textit{missing}~\cite{bareinboim2016causal}. To fill in the missing answer, current works~\cite{xu2019end,wei2018task} proposed to build patient simulators on the collected data. In specificity, they adopt patient-doctor conversation records~\cite{schatzmann2007agenda,li2016user} to generate responses, and use the default answer to respond with the counterfactual inquiry, which will introduce unexpected collider/selection biases. 
The collider/selection biases are introduced when there are controlled colliders~\cite{cole2010illustrating}. These biases usually cannot be solved by boosting with more data~\cite{pearl2009causality}. The most straightforward solution is to develop a new simulator that does not respond with the counterfactual inquiry using the default answer but sampling the answer from the authentic distribution. To model the answer distribution from the collected data, ~\cite{peng2018deep} incorporated a parametric model of the environment into the dialogue agent to generate a simulated user experience, which however is prone to overfit the characteristics of the training data~\cite{chen2017survey}. Instead of direct estimation, matching techniques have been shown to be better at producing less biased and more causality-consistent estimation~\cite{stuart2010matching}. In MDA, the collected data are usually very sparse therefore ordinary matching methods are usually infeasible due to the problem of \textit{lack of overlap}. To resolve this problem, propensity score matching~(PSM)~\cite{dehejia2002propensity,rosenbaum1983central} is introduced to estimate the low-dimension propensity of the instance for matching. However, PSM requires a reliable propensity score estimator to perform well. In the sequential interaction process, the covariates will be affected by the previous unfamiliar action which might degrade the reliability of the estimated propensity score. Our PLM is not the first method to introduce latent features for improving propensity score matching~\cite{jakubowski2015latent}, which limits the measurement error of propensity score. Different from prior works, the latent features in PLM are not given at the beginning and are learned through self-supervised learning. Moreover, PLM directly matches the latent features since the propensity score is affected by the inquiry. Similar to our PLM, a recent study in self-supervised learning, masked auto-encoder (MAE)~\cite{he2022masked} also observe that latent learning generally has better generalization ability. Differently, our PLM is optimized to model the action distribution while MAE is optimized for general purposes and requires extra task-specific tuning for downstream applications.

As for the task-oriented dialogue agent, most of the current task-oriented dialogue systems adopt the framework of reinforcement learning~(RL)~\cite{mnih2015human,lipton2017bbq,li2017end}, and some works~\cite{madotto2018mem2seq,wu2019global,lei2018sequicity} adopt the sequence-to-sequence style for dialogue generation. For medical dialogue systems, due to a large number of symptoms, reinforcement learning is a better choice for topic selection~\cite{tang2016inquire,kao2018context,peng2018refuel}. In the context of MDA, the actions of symptom inquiry and disease diagnosis are discrete. Therefore, most of the current MDA methods exploit the classical discrete control methods, e.g., Deep Q-Network~(DQN)~\cite{mnih2015human} to select the actions. ~\cite{tang2016inquire} applied DQN to diagnose using synthetic data. While ~\cite{wei2018task} first did experiments on real-world data using DQN. To include explicit medical inductive bias for improving the diagnostic performance, KR-DQN~\cite{xu2019end} {proposed} an end-to-end model guided by symptom-disease knowledge priors. KR-DQN applies the predefined conditional probability of symptom and disease to transform the estimated Q-values.

Differently, our work follows the logic of MDA in real life and formulates it as a sequential discriminative decision-making problem. To model the decision-making process, neuroeconomics proposes a diffusion model~\cite{10.1162/neco.2008.12-06-420}. 
Inspired by these discoveries, our proposed dual-process P2A also incorporates the concept of decision threshold for the diagnoser to determine when to inform disease. Like ours, PG-MI-GAN~\cite{xia2020generative} adopts a separate diagnoser from the inquiry policy. Specifically, it trains an inquiry generator to generate a sequence of inquiries undistinguished by the discriminator and further uses a pretrained diagnoser to finetune the generator.
As we discovered, the design of the diagnoser needs to take the indirect bias~\cite{vanderweele2013three}, distributional bias~\cite{pearl1995causal,koller2009probabilistic,huang2012pearl,shpitser2012identification} as well as training bias~\cite{mcallister2018robustness} into account to better capture the symptom-disease relationship and improve the \textit{transportability}~\cite{pearl2014external} of the diagnosis agent.

Moreover, as a medical application, an MDA agent is also required to consider the uncertainty of its decision to provide a robust and trustful diagnosis result for the ethics concern. Most of the current MDA agent is allowed to jump into informing a disease without any regulation. Instead, our MDA agent takes the uncertainty to augment the decision-making process. There are plenty of works studying how to combine uncertainty and exploration~\cite{whaite1997autonomous, whaite1991uncertainty, mcallister2018robustness, osband2018randomized, burda2018exploration,brunke2022safe}. However, different from ours, these works do not employ uncertainty to provide a stop mechanism for sequential decision-making.

\section{Preliminaries}

\begin{figure*}[t]
	\centering
	\includegraphics[width=\linewidth]{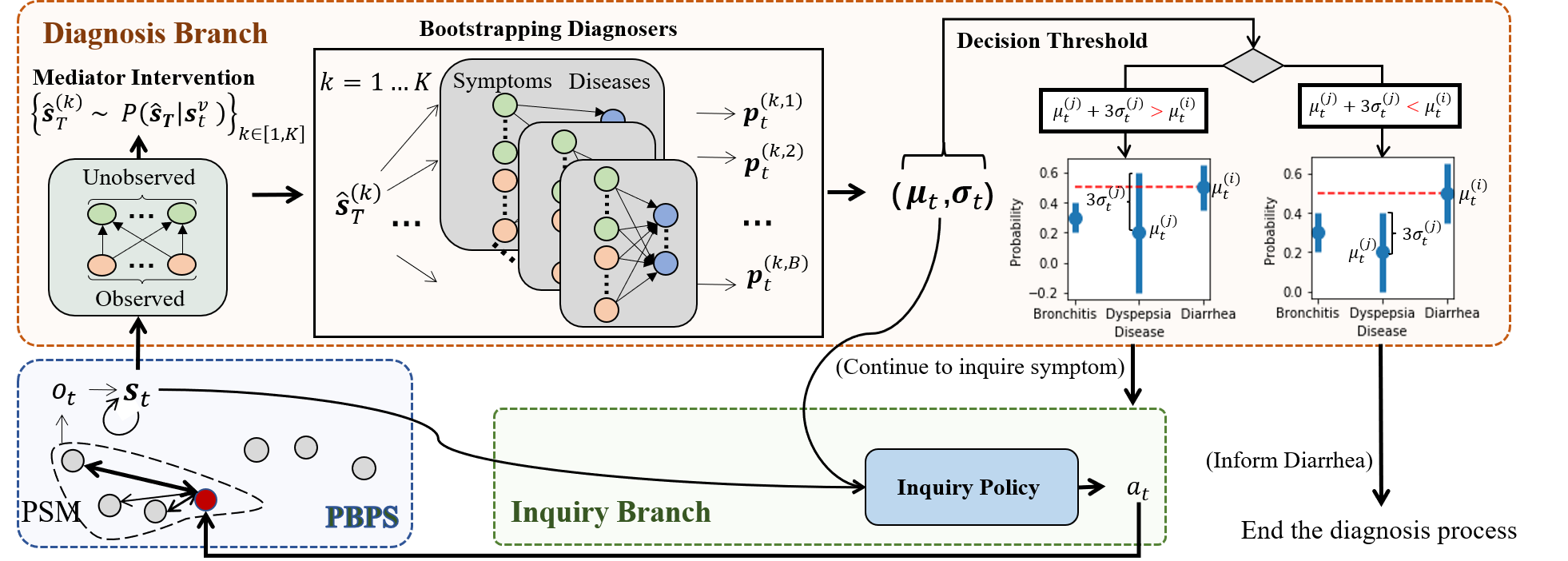}
	 			\vspace{-6mm}
	\caption{An overview of our CA-MDA. The diagnosis branch takes $\mathbf{s}_t$ to draw $K$ final states which are fed into the $B$ bootstrapping diagnosers to obtain the expectation and deviation. P2A keeps inquiring about symptoms of the patient simulator until the diagnosis meets the decision threshold.}
	\vspace{-6mm}
	\label{fig:pipeline}
\end{figure*}

\textbf{Medical diagnosis assistant (MDA).} As shown in Fig.~\ref{fig:cfd}, a diagnosis record consists of a self-report, symptom statuses $\mathbf{y} = [y_a]_{a=1}^{N_s} \in \{-1, 0, 1\}^{N_s}$~(-1 for `no', 0 for `not sure', 1 for `yes'), and a ground-truth disease label $d \in [1, ..., N_{d}]$, where ${N_s}$ is the number of symptoms and ${N_d}$ is the number of diseases. An MDA diagnosis process starts with a patient's self-report, which forms the initial dialogue state, $\mathbf{s}_{t=0} \in \mathcal{S}$, where $\mathcal{S} \subseteq \mathbb{R}^{N_s}$ denotes the state space and $\mathbf{s}_t$ maintains the values of all mentioned symptoms (i.e., -1 for `no', 0 for `not sure' and 1 for `yes') up to timestep $t$. And $\mathcal{A} = [1, ..., {N_s}, {N_s}+1, ..., {N_s}+{N_d}]$ represents the action space of the agent. The MDA agent selects an action $a_t \in \mathcal{A}$, to either inquire about symptoms ($a_t \le N_s$) or inform the diagnostic result ($a_t > N_s$). And since the interaction ends when the discriminative action is made, we use $T$ to denote the timestep when the diagnosis ends and $a_T$ stands for the diagnosis action/decision and $\mathbf{s}_T$ is the final state of the diagnosing process. The patient simulator updates the dialogue state by $\mathcal{P}: \mathcal{S} \times \mathcal{A} \rightarrow \mathcal{S}$. Practically, the patient simulator will return a response/observation $o_t$ to the inquiry $a_t$, and then update the dialogue state according to $o_t$. Although the observation $o_t$ is assumed to have the same value as the symptom status $y_{a_t}$ in MDA, we will use them differently according to the contexts. To indicate whether a symptom has been inquired about, we use a binary vector $\mathbf{m}_t$ to record the visited inquiries up to time $t$, where $m_{t, a_{t' \le t}} = 1$. MDA is a sequential discriminant problem that sequentially inquires about symptoms and makes a discriminative diagnosis at the end. Formally, MDA is to search for an agent $\pi: \mathcal{S} \rightarrow \mathcal{A}$ that:
\begin{align}
    \max_{\pi} \mathbb{E}_{\mathcal{P}, \pi}\big[-\max (T, T^*) +  \mathbb{I}(a_T = N_s+d) \big],
    \label{equ:sqd}
\end{align}
where $T^*$ stands for the optimal number of inquiries for patient $\mathcal{P}$. For different patients, $T^*$ varies as the inquiries required for the diagnosis varies. $\mathbb{I}(\cdot)$ is an indicator function that returns $1$ if the propositional logic formula in the bracket is satisfied, otherwise returns $0$. The optimality of Equ. (\ref{equ:sqd}) is obtained when the inquiry process collects just enough symptom information for an accurate diagnosis. To this end, there exists a subordinate relationship among the actions, i.e., inquiry serves for discrimination.

\textbf{Reinforcement learning (RL) for MDA.}  To apply RL for resolving MDA, previous methods~\cite{xu2019end,wei2018task} design complex reward functions, i.e., $\mathcal{R}: \mathcal{S} \times \mathcal{A} \rightarrow \mathbb{R}$. for different inquiry/diagnosis actions. The target of reinforcement learning is to solve via a policy form $\pi: \mathcal{S} \rightarrow \mathcal{A}$, which maximizes the expected sum of rewards:
\begin{align}
\max_{\pi} \mathbb{E}\Big[\sum_{t=0, a_t\sim\pi(\mathbf{s}_t)}^T\gamma^t\mathcal{R}(\mathbf{s}_t, a_t)\Big], \label{equ:rl}
\end{align}
where $\gamma \in [0, 1)$ is a discount factor. However, since all actions are abstracted into numerical rewards and connected by accumulation in Equ. (\ref{equ:rl}), the subordinate relationship between inquiry and diagnosis cannot be modeled explicitly. To mitigate this issue, previous methods assign significantly larger rewards/punishments for correct/incorrect diagnoses than inquiries. This strategy is based on the numerical association between inquiry and diagnosis, instead of the causal relationship. Inference based on association usually requires a dedicated design for reward functions to balance the importance of different actions. And it would also cause some unexpected behavior when handling out-of-distribution cases where the association is different, like making a rushy decision without inquiring~\cite{xu2019end,wei2018task}.

\begin{figure}[!t]
    \centering
    \includegraphics[clip=True, trim={0 10pt 0 0}, width=\linewidth]{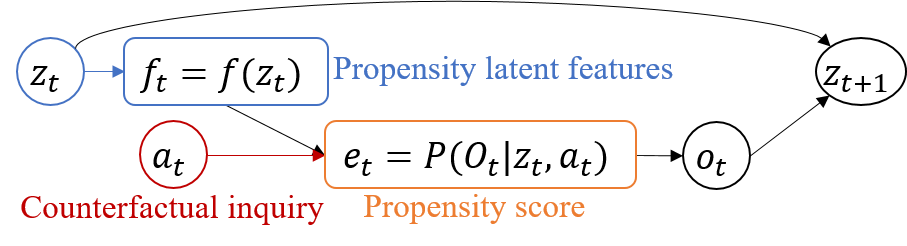}
    \vspace{-5mm}
    \caption{Diagram illustration of propensity latent features. $z_t$ is the state of the simulator and $a_t$ is the inquiry and $o_t$ is the answer. The propensity score $e_t$ is affected by $a_t$ while the propensity latent features $f_t$ are not.}
    \label{fig:propensity_feature}
\end{figure}

\section{Causality-Aware MDA}
To mitigate the two identified biases that exist in previous RL-based methods, i.e., the default-answer bias and the distributional inquiry bias, we propose our CA-MDA, which is comprised of the propensity-based patient simulator~(PBPS) and the progressive assurance agent~(P2A). Moreover, P2A leverages a neuro-inspired decision threshold to establish the subordinate relationship between inquiry and diagnosis explicitly. The overview of the interaction between PBPS and P2A is shown in Fig.~\ref{fig:pipeline}.

\subsection{Propensity-based Patient Simulator}

As shown in Fig.~\ref{fig:propensity_feature}, a simulator is required to generate a counterfactual response $o_t$ to a counterfactual inquiry $a_t$ according to the simulator state $z_t=(\mathbf{y}_t, d)$. Different from the dialogue state $\mathbf{s}_t$, simulator state $z_t$ includes the symptom statuses $\mathbf{y}_t$ and the disease label $d$. And $\mathbf{y}_{t=0}$ is the original symptom existences  $\mathbf{y}$ of the anchor record. $\mathbf{y}_t$ is updated after inferring an unrecorded symptom existence. To causally predict the answer $o_t$, one of the most popular causal inference algorithms is propensity score matching~(PSM). Specifically, PSM conducts matching based on the sample's propensity score $e_t=P(O_t|z_t, a_t)$. 
After matching, PSM refers to the observation from the matched samples as a potential answer. By drawing the answers from matched samples, the generated answers of PSM are more authentic and less biased. However, the propensity score is conditioned on the inquiry as shown in Fig.~\ref{fig:propensity_feature}. When $a_t$ is a counterfactual inquiry, the combination of ($z_t, a_t$) might be unfamiliar to the propensity estimators and cause the estimation unreliable. As the interaction goes on, the error will be accumulated, forming a vicious spiral.

\textbf{Propensity latent matching}. To this end, we propose a novel matching strategy, Propensity Latent Matching~(PLM). Different from the propensity score, our method takes a step back and matches with the propensity of latent features $f_t=f(z_t)$  before the inquiry takes effect, where
\begin{align}
 e_t = P(O_t|z_t, a_t)=P(O_t|f(z_{t}), a_t).
 \label{equ:plm}
\end{align}
Equ.(\ref{equ:plm}) shows that PLM is mathematically equivalent to PSM. The relationship between $f_t$ and $e_t$ is shown in Fig.~\ref{fig:propensity_feature}. From the illustration, we can see that the benefit of PLM is that $f_t$ is not affected by the counterfactual inquiry $a_t$, and thus the counterfactual inference is more reliable. And if $a_t$ is still unobserved among the matched records, $o_t$ will be 0, and $z_{t+1}$ remains the same as $z_{t}$. This prevents the simulator state from becoming out-of-distribution during the interaction process.

In building PBPS, we use multilayer perceptron~(MLP) $f_{\phi_{\rm L}}(\cdot)$ to model propensity latent features $f_t$, where ${\phi_{\rm L}}$ denotes the parameter of the network. And we use another MLP $e_{\phi_{\rm S}}(f_{\phi_{\rm L}}(\cdot), a)$ parameterized by $\phi_{\rm S}$ to model the propensity score $e_t$ satisfying Equ.(\ref{equ:plm}). To optimize $\phi_{\rm L}$ and $\phi_{\rm S}$, we deploy the self-supervised strategy. Specifically, we treat the recorded symptom existence $\mathbf{y}$ as the $\mathbf{y}_T$. To obtain the immediate simulator state (e.g., $z_t$ in Fig.~\ref{fig:propensity_feature}), we mask off some of $\mathbf{y}$ according to the visitation mask $\mathbf{m}_t$. The mask is used to mask off the symptoms that are mentioned after timestep $t$ in the diagnosis record, resulting in the immediate state $z_t = (\mathbf{y}\odot \mathbf{m}_t, d)$. To this end, $e_{\phi_{\rm S}}(f_{\phi_{\rm L}}(\cdot), a)$ can be learned via minimizing the cross entropy loss CE:
\begin{equation}
\label{equ:prop}
\underset{\phi_{\rm L},\phi_{\rm S}}{\min} \mathbb{E}_{\mathbf{y}, d, \mathbf{m}_t, a} \Big[\text{CE}\Big(e_{\phi_{\rm S}}\big(f_{\phi_{\rm L}}(\mathbf{y} \odot \mathbf{m}_t, d), a\big), y_a\Big) \Big],
\end{equation}
where $a$ is the index of one of the masked symptoms and $y_a \neq 0$ is the symptom existence of the $a$-th symptom. After training, the output of $f_{\phi_{\rm L}}(\cdot)$ is the propensity latent features for matching.

During inference, for a counterfactual inquiry $a_t$, the simulator explores all records $\{q_i\}_{i=1:N}$ that have the same disease (i.e., $d^{(q)} = d$) and also have $a_t$ in their observed symptoms (i.e., $y^{(q)}_{a_t} \neq 0$). The similarity weight of a record $q$ is formulated as 
\begin{small}
	\begin{equation}
	\begin{aligned}
	\label{equ:ret}
	P( q | f_t, a_t ) \propto \mathbb{I}\Big((d^{(q)} =  d)\land (y^{(q)}_{a_t} \neq 0)\Big) \times e^{-\| f_t-f^{(q)} \|^2 / \sigma^2},
	\end{aligned}
	\end{equation}
\end{small}
where $f^{(q)}=f_{\phi_{\rm L}}(\mathbf{y}^{(q)}\odot \mathbf{m}_t, d^{(q)})$ and $e^{-\| f_t-f^{(q)} \|^2 / \sigma^2}$ is a bell-shaped function for normalizing the similarity into probability ($\sigma$ indicates a standard deviation). Here, $\mathbf{m}_t$ indicates the visited symptoms of $\mathbf{y}_t$. The similarity of propensity latent features of the patient records implies that the existences of symptoms of these records are more probably similar. Then the patient simulator can sample a record according to the similarity weight, i.e., $q^\prime \sim P( q | f_t, a_t)$. And the symptom existence $y^{(q^\prime)}_{a_t}$ of the sample record is used as the answer to the inquiry $a_t$. The state transition of PBPS with anchor record $p$, i.e., $\mathbf{s}_{t+1} \leftarrow \mathcal{P}_{\rm PBPS}(\mathbf{s}_t, a_t; p)$, is presented in Alg.~\ref{alg:pipeline_usersim}, where line 7 (highlighted in bold and italics) is the extra operation of our simulator compared to the previous simulator $\mathcal{P}_{\rm PS}$. 
\begin{algorithm}[!t]
	\small
	\begin{algorithmic}[1]
		\renewcommand{\algorithmicrequire}{\textbf{Input:}}
		\renewcommand{\algorithmicensure}{\textbf{Output:}}
\Require $\mathbf{s}_t, a_t$, simulator state $z_t$ and the self-report
		\Ensure $\mathbf{s}_{t+1}$
		\If {t = -1}
		\State Initialize the dialogue state $\mathbf{s}_0 \leftarrow [0, ..., 0]$ and the visitation indicator $\mathbf{m}_0 \leftarrow [0, ..., 0]$ with the size of $N_s$, and initialize the simulator state $z_0 \leftarrow (\mathbf{y}, d)$
		\State Let $s_{0, a} \leftarrow y_a$ and $m_{0, a} \leftarrow 1$ for each $a$ parsed from the self-report
		\State Return $\mathbf{s}_0$
		\Else
		\State Initialize $a \leftarrow a_t, \mathbf{s}_{t+1} \leftarrow \mathbf{s}_t, \mathbf{m}_{t+1} \leftarrow \mathbf{m}_t, z_{t+1} \leftarrow z_t, q^\prime \leftarrow p$
		\State \textit{\textbf{if $\big ( (y_a = 0) \land (m_{t+1, a} = 0)\big )$, sample $q^\prime \sim P( q | f_t, a)$ according to Equ.(\ref{equ:ret}).}}
            \State Generate response $o_t \leftarrow y^{(q^\prime)}_a$
		\State Update dialogue state $\mathbf{s}_{t+1}$ by $s_{t+1, a} \leftarrow o_t$
        \State Update visitation history $\mathbf{m}_{t+1}$ by $m_{t+1, a} \leftarrow 1$
        \State Update simulator state $z_{t+1}$ by $y_{t+1, a}\leftarrow o_t$
		\State Return $\mathbf{s}_{t+1}$
		\EndIf
	\end{algorithmic} 
	\caption{\textbf{Propensity-based patient simulator~(PBPS)}: $\mathcal{P}_{\rm PBPS}(\mathbf{s}_t, a_t; p)$}
	\label{alg:pipeline_usersim}
\end{algorithm}

\subsection{Progressive Assurance Agent}

Our \emph{Progressive Assurance Agent} (P2A) consists of two separate yet cooperative branches for symptom inquiry and disease diagnosis, as shown in Fig.~\ref{fig:pipeline}. The inquiry branch inquires about symptoms to get $\mathbf{s}_t$ from PBPS to increase diagnosis confidence, while the diagnosis branch imagines and reasons the future scenarios to robustly estimate the disease and its confidence per step until the confidence is high enough~(satisfying \textit{Decision Threshold} (DT)~\cite{ratcliff2016diffusion}) to inform a disease. In this manner, if the patient is unfamiliar/out-of-distribution to the agent, the decision threshold will modulate the inquiry branch to adapt to the unfamiliar case. 
Formally, the overall optimization target of P2A is:
\begin{align}
    \max_{\pi_\theta, f_\phi} \mathbb{E}_{\mathcal{P}, \pi_\theta}\Big[\sum_{t=0}^{\max (T, T_\text{DT})-1} \gamma^t r + \mathbb{I}\big(f_{\phi}(\mathbf{s}_{T_\text{DT}}) = d\big) \Big], \label{equ:p2a}
\end{align}
where $T_\text{DT}$ stands for the \textit{minimal number of inquiries for reaching the decision threshold (DT)}, $r$ is a negative constant, and $\mathbf{s}_{T_\text{DT}}$ stands for the state when the decision threshold is met. $\pi_\theta$ is the inquiry policy parameterized by $\theta$ and $f_{\phi}$ denotes the diagnoser parameterized by $\phi$. Different from the RL target in Equ. (\ref{equ:rl}), our P2A explicitly models the subordinate relationship between inquiry and diagnosis via the decision threshold as formulated in Equ. (\ref{equ:p2a}). Moreover, after introducing the decision threshold, the inquiry policy and the diagnostic agent can be optimized separately. The left-hand side is the optimization target for the inquiry branch, i.e., to reach the decision threshold as soon as possible. The right-hand side is the target for the diagnosis branch, i.e., to accurately discriminate disease when the decision threshold is met. In the following, we will elaborate on the formulation of the diagnosis branch and inquiry branch. The algorithm pipeline of P2A is illustrated in Alg.~\ref{alg:pipeline}

\textbf{Diagnosis Branch:} 
\label{sec:mediator}
To eliminate the distributional inquiry bias on diagnosis, our P2A intervenes in the ``non-inquired observation" to be independent of the inquiry behavior. Specifically, our P2A first infers the possible final state $\mathbf{\hat{s}}_T$ (i.e., the state at the end of the diagnosing process) according to the inquired observation $\mathbf{s}_t^v$, i.e., $P(\mathbf{\hat{s}}_T | \mathbf{s}_t^v)$. Then, our P2A intervenes in the observations of the non-inquired symptoms $\mathbf{s}_t^u$ with the estimated simulated patient, i.e., do($\mathbf{s}_t^u = \mathbf{\hat{s}}_T - \mathbf{s}_t^v$). One can intuitively treat such intervention processes as ``imagining and reasoning the future interactions". After that, our P2A feeds the inquired and intervened symptoms as input to the discriminative diagnoser $P(d|\hat{\mathbf{s}}_T)$. In total, the diagnoser of our P2A diagnoses by using the observed symptoms via $P(d|\mathbf{s}^v_t)=\sum_{\hat{\mathbf{s}}_T}P(d|\hat{\mathbf{s}}_T)P(\hat{\mathbf{s}}_T|\mathbf{s}^v_t)$. We train the intervener $f_{\phi_{\rm G}}(\cdot)$ and the  discriminative diagnoser $f_{\phi_{\rm B}}(\cdot)$ for $P(\mathbf{\hat{s}}_T | \mathbf{s}_t^v)$ and $P(d | \mathbf{\hat{s}}_T)$, respectively.

\begin{algorithm}[t]
\small
\begin{algorithmic}[2]
   \renewcommand{\algorithmicrequire}{\textbf{Input:}}
    \renewcommand{\algorithmicensure}{\textbf{Output:}}
    \Require Initial inquiry policy parameters $\theta$, intervener parameters $\phi_G$, $B$ bootstrapping diagnosers parameters $\phi_{\rm B}$, empty replay buffer $\mathcal{D}_Q$ and final state buffer $\mathcal{D}_C$.
    \For {each episode}
        \State Get one anchor record $p$ and initialize the state $\mathbf{s}_0$
        \For {t = 0 : $T_\text{max}$}
        \State Use the intervener to generate $K$ final states: $\{\mathbf{\hat{s}}_T^{(k)} \sim f_{\phi_G}(\mathbf{s}_t), k \in [1, K]\}$
        \State Feed the final states to $B$ bootstrapping diagnosers: $\mathbb{P}_t= \{\mathbf{p}_t^{(k, b)}, k \in [1, K], b \in [1, B]\}$
        \State Calculate the statistics $\boldsymbol{\mu}_t, \boldsymbol{\sigma}_t$ of $\mathbb{P}_t$
        \If {DT($\boldsymbol{\mu}_t, \boldsymbol{\sigma}_t$) $\vee (t = T_\text{max})$} \Comment{Meet DT or time out}
        \State Inform disease: $ a_t = (\text{max}_{i}\mu_{t}^{(i)})+N_s$
        \ElsIf{10\% chance} \Comment{$\epsilon$-greedy explore}
        \State $a_t$: inquire about a random unobserved symptom
        \Else \Comment{Use inquiry policy to select inquiry}
            \State $a_t = \text{max}_{a} Q_\theta(\mathbf{s}_t, \boldsymbol{\mu}_t, a)$
        \EndIf
        \State Interact with $\mathcal{P}_{\rm PBPS}$: $\mathbf{s}_{t+1} = \mathcal{P}_{\rm PBPS}(\mathbf{s}_t, a_t; p)$
        \State Store transition ($\mathbf{s}_{t-1}$, $\boldsymbol{\mu}_{t-1}$, $a_{t-1}$, $\mathbf{s}_{t}$, $\boldsymbol{\mu}_{t}$) in $\mathcal{D}_{\text{Q}}$
        \If {$a_t > N_s$} \Comment{Action of informing disease}
        \State Store records (${\mathbf{s}}_{T=t}, d^{(p)}$) in $\mathcal{D}_{\text{C}}$
        \State \textbf{break}
        \EndIf
        \If {time to update}
        \State Sample mini-batch from $\mathcal{D}_{Q}$ and update $\theta$ (Equ.(\ref{equ:updateq}) and Equ.(\ref{equ:updateqt})).
        \State Sample $B$ mini-batches from $\mathcal{D}_{C}$ with replacement to update $\phi_{\rm G}$ and $\phi_{\rm B}$ (Equ.(\ref{equ:gene}) and Equ.(\ref{equ:bootstrap})).
         \EndIf
\EndFor
\EndFor
	\end{algorithmic} 
	\caption{\setlength{\tabcolsep}{2.5pt}
		\textbf{Progressive Assurance Agent~(P2A)}}
	\label{alg:pipeline}
\end{algorithm}

\textit{Intervener} aims at predicting the final symptom state $\hat{\mathbf{s}}_T$ from the current inquired state $\mathbf{s}_t^v$. 	
Therefore, we model it as a generative problem. $\phi_{\rm G}$ is the parameter of the generator $f_{\phi_{\rm G}}(\cdot)$, whose target is:
\begin{equation}
\label{equ:gene}
\underset{\phi_{\rm G}}{\min} \mathbb{E}_{\mathbf{s}_T \sim \mathcal{D}_c}\Big[\text{CE}\Big(f_{\phi_{\rm G}}(\mathbf{s}_T \odot \mathbf{m}_t)_a, s_{a}\Big)\Big],
\end{equation}
where $\mathbf{m}_t$ is the visitation mask indicating which symptoms are observed up to time $t$, $a$ is the index of a masked symptom, i.e., $m_{t, a}=0 \land m_{T, a}=1$. The target for $f_{\phi_{\rm G}}$ is to recover the masked information. And the final state $\mathbf{s}_T$ is sampled from the data replay buffer $\mathcal{D}_c$ which stores the final state during the training process (line 16 in Alg.~\ref{alg:pipeline}).  As shown in the overview~(Fig. \ref{fig:pipeline}), the Monte Carlo sampling is applied by obeying the generative model $f_{\phi_{\rm G}}(\mathbf{s}_t)$ to sample $K$ possible final states $\{\hat{\mathbf{s}}_T^{(k)}\}_{k=1}^{K}$. Note that, the inquired symptoms of $\mathbf{s}_t^v$ remain the same in these final states. After the intervention, the $K$ imaginary final states are fed to the diagnoser $f_{\phi_{\rm B}}(\cdot)$ to check whether the decision threshold is met or not.

\par{\textit{Decision Threshold.}} Intuitively, doctors stop inquiring to inform diseases when they are \emph{confident} that inquiring about more symptoms would not overturn his diagnosis. Therefore, we propose the decision threshold~(DT) to mimic such an introspective process, that is, the agent would stop inquiring to inform the preferred disease \textit{if the agent believes that the probability of the preferred disease is high enough so that inquiring more symptoms would not overturn the preferred disease probabilistically.} To estimate the probability of each disease and its confidence, bootstrapping technique~\cite{whaite1991uncertainty,whaite1997autonomous,mcallister2018robustness} is adopted to train ensembles of diagnosers.

\par{Bootstrapping diagnosers} are trained to diagnose using the final state $\mathbf{s}_T$ stored in the data replay buffer. The target of diagnoser $i$ with parameter $\phi_{\rm B, i}$ is
\begin{equation}  
\label{equ:bootstrap}
\min_{\phi_{\rm B, i}} \mathbb{E}_{(\mathbf{s}_T, d) \sim \mathcal{D}_c} \Big[\text{CE}(f_{\phi_{\rm B, i}}(\mathbf{s}_T), d)\Big].
\end{equation}
Note that, since P2A terminates the interaction when the decision threshold is met, $\mathbf{s}_T$ can be regarded as the same as $\mathbf{s}_{T_\text{DT}}$. In this sense, Equ. (\ref{equ:bootstrap}) is optimized for the right-hand side of Equ. (\ref{equ:p2a}).

During inference, the final states sampled from the intervener are fed into $B$ bootstrapping diagnosers, resulting in a final disease probability set $\{\mathbf{p}_t^{(k,b)}\}^{K,B}_{k=1, b=1}$. The final disease probability set is then used to calculate the expectation $\boldsymbol{\mu}_t=[{\mu}^{(1)}_t,\cdots,{\mu}^{(N_d)}_t]$ and standard deviation $\boldsymbol{\sigma}_t=[{\sigma}^{(1)}_t,\cdots,{\sigma}^{(N_d)}_t]$ of diseases: 
\begin{small}
	\begin{equation} 
	\begin{aligned}
	\label{equ:stati}
	\boldsymbol{\mu}_t = \frac{1}{KB} \sum_{b=1}^{B}\sum_{k=1}^{K} \mathbf{p}_t^{(k,b)}, 
	\quad 
	\boldsymbol{\sigma}_t^2 = \frac{1}{KB} \sum_{b=1}^{B}\sum_{k=1}^{K} (\mathbf{p}_t^{(k,b)} - \boldsymbol{\mu}_t)^2,
	\end{aligned}
	\end{equation}
\end{small}which are further used to calculate the confidence intervals of diseases. Moreover, bootstrapping diagnosers are also popular in reducing the unexpected noisy effect introduced by the data sampling process and parameter initialization. This kind of noisy effect would hamper the model's in-distributional performance~\cite{mcallister2018robustness}.

With the mean and standard deviation, DT would be met if the probability of the preferred disease is beyond the upper bound of the 6$\sigma$ confidence interval~\cite{pande2001six,parzen1960modern} of the other diseases' probabilities. Denote the preferred disease as $i$, i.e., $i = \text{argmax}_j  \mu^{(j)}_t, \forall j \in [1, ..., N_d]$. DT is formulated as

\begin{small}
	\begin{equation}
	\label{equ:stop}
	\text{DT}(\boldsymbol{\mu}_t, \boldsymbol{\sigma}_t)=
	\begin{cases}
	\text{True}, &  \forall j \neq i, \mu_t^{(i)} > \mu_t^{(j)} + 3\sigma_t^{(j)},   \\
	\text{False}, &  \text{otherwise}
	\end{cases}.
	\end{equation}
\end{small}

\textbf{Inquiry Branch:} The left-hand side of Equ.(\ref{equ:p2a}) is the problem of finding the shortest path to reach the decision threshold, which can be well-addressed by reinforcement learning. Therefore, we apply deep Q-learning~\cite{mnih2015human,wang2015dueling} to optimize the inquiry policy. Specifically, the inquiry branch is modeled by a $Q$ network~\cite{wang2015dueling} parameterized by $\theta$, which takes the concatenation of the state $\mathbf{s}_t$ and the current disease probabilities $\mathbf{u}_t$ to predict the inquiry action $a_t \in [1, ..., N_s]$.
\begin{equation}
\label{equ:policy}
a_t = \text{max}_a Q_\theta(\mathbf{s}_t, \mathbf{u}_t, a).
\end{equation} The optimization target of inquiry policy is:
\begin{small}
	\begin{equation}
	\begin{aligned}
	\label{equ:updateq}
	\min_{\theta} \mathbb{E} \Big [ \big|\big| r+\gamma \text{max}_{a} Q_{\theta_{\text{targ}}}(\mathbf{s}_{t+1}, \mathbf{u}_{t+1}, a) - Q_\theta(\mathbf{s}_t, \mathbf{u}_t, a_t) \big|\big|_2^2 \Big ],
	\end{aligned}
	\end{equation}
\end{small}
where $r=-0.1$ is a negative constant for punishing redundant inquiries. And the training data are sampled from the replay buffer $\mathcal{D}_Q$ (line 14 in Alg.~\ref{alg:pipeline}). The parameter $\theta_{targ}$ is used for stabilizing the training, updated with a momentum factor $\alpha$:
\begin{equation}
\label{equ:updateqt}
\theta_{\text{targ}} = \alpha \theta_{\text{targ}} + (1-\alpha) \theta.
\end{equation}

\section{Experiments}
To justify the effectiveness of our CA-MDA, we mainly focus on answering the following questions: \textbf{1)} \emph{Is the propensity-based patient simulator better at answering counterfactual symptom inquiries? (Reducing default-answer bias)} \textbf{2)} \emph{Does the progressive assurance agent achieve better in-distribution and out-of-distribution diagnostic performance? (Reducing distributional inquiry bias)} \textbf{3)} \emph{Does the decision threshold mechanism bring about a more robust diagnosing process?} More converging curves, hyperparameter analysis, and extended experiments are presented in Appendix C. For better understanding, we also provide numerical running examples in Appendix D.

\textbf{Dataset.} We perform extensive and comprehensive evaluations on two MDA benchmarks, \emph{i.e.}, MuZhi~(\textbf{MZ}) \cite{wei2018task} composed of 586 training and 142 test records with 66 symptoms and 4 diseases; DingXiang~(\textbf{DX}) \cite{xu2019end} composed of 423 training and 104 test records with 41 symptoms and 5 diseases. More details about the benchmarks are placed in Appendix C.
$\mathcal{P}^{\rm train}$ and $\mathcal{P}^{\rm all}$ denote the patient simulators organized by the training records and all records~(for training and testing) respectively. For instance, $\mathcal{P}^{\rm train}_{\rm PBPS}$ represents the PBPS using training records in a benchmark to interact with the agent.  

\par{\textbf{Baselines.}} To answer Question 1), we have compared our PBPS with the previous patient simulator (PS)~\cite{wei2018task,xu2019end}. To our knowledge, all prior MDA works did not propose new patient simulators but used PS with the default-answer strategy. As for the direct estimator, we adopt a generative world model~(GEN)~\cite{peng2018deep}. {PBPS-PSM is our propensity-based simulator using propensity score matching and PBPS uses our propensity latent matching.} As for Question 2), we have compared our P2A against three baselines using our PBPS, i.e., DQN~\cite{wei2018task}, KR-DQN~\cite{xu2019end}, and PG-MI-GAN~\cite{xia2020generative}. DQN combines symptom inquiry and disease diagnosis into a single policy network and trains it by deep Q-learning~\cite{mnih2015human,mnih2013playing}. Improved from DQN, KR-DQN~\cite{xu2019end} adds a knowledge-routing module at the head of the policy using predefined disease-symptom knowledge, i.e., matrices of conditional/joint probability. Distinguished from DQN and KR-DQN, our P2A disentangles the disease diagnosis from the policy. Similar to ours, PG-MI-GAN~\cite{xia2020generative} adopts a separate diagnoser from the inquiry policy. Specifically, it trains an inquiry generator to generate a sequence of inquiries that are hard for the discriminator to distinguish, and further uses a pretrained diagnoser to finetune the generator. Its inquiry generator is pretrained through imitation, which lacks reasoning between inquiries and diseases. Moreover, its diagnoser is fixed during the finetuning process and does not evolve with the inquiry generator. To testify to the robustness and safeness of the decision threshold mechanism (i.e., Question 3), we examine whether the accuracy of P2A can be improved significantly when the decision threshold is met during diagnosing under both \textbf{In.} and \textbf{Out.} settings.

\subsection{Evaluation on PBPS}

To evaluate how accurately a simulator can answer the counterfactual symptom inquiries, we propose a novel casual metric, named \emph{Inference Ability} (IA).

\textbf{Inferential ability}. 
Factually, there is no ground truth for the counterfactual symptom inquiries. In MDA, fortunately, we know what should have been observed if the inquired symptom was not inquired about, that’s, `not sure', as shown in Fig.~\ref{fig:collider}. This allows us to convert some recorded symptoms to be unrecorded and relabel their observation as `not sure' to synthesize new records. Building simulators with the relabeled records, we can examine how the simulators answer the counterfactual symptom inquiries by inquiring about the relabeled symptom inquiries. After that, we can use the original observations to compare with the answers from the simulators. 
\begin{table}[!t]
	\centering
	\setlength{\tabcolsep}{3pt}
	\caption{The quantitative evaluation of the patient simulators by using the Inferential Ability~(IA) metric.}
	\begin{tabular}{c|ccccc} \toprule
		Benchmarks             & PS~\cite{xu2019end,wei2018task}  & GEN~\cite{peng2018deep}  & PBPS-PSM & PBPS*               & PBPS             \\ \midrule
		MZ & 0.0\textpm 0.0 & 0.12\textpm 0.01 & 0.23\textpm 0.03 & 0.57\textpm 0.02  & \textbf{0.62\textpm 0.02} \\ 
		DX & 0.0\textpm 0.0 & 0.14\textpm 0.03 & 0.27\textpm 0.03 & 0.58\textpm 0.03 & \textbf{0.66\textpm 0.01}                          \\ \bottomrule
	\end{tabular}
	\label{tab:ia}
\end{table}

Practically, to measure the IA of the patient simulators, we relabel all \textit{implicit symptoms} from the records and only provide the record $p_\text{ex}$ with explicit symptoms and disease to the patient simulators. After that, we inquire about the relabeled implicit symptom of different patient simulators and calculate the accuracy of their responses. Formally,
\begin{equation}
{\rm IA} = \frac{1}{D}\sum_{p \in {\rm test set}}\frac{1}{|\mathbf{y}_\text{im}|}\sum_{y_a \in \mathbf{y}_\text{im}}\mathbb{I}(\mathcal{P}^{\rm train}(\mathbf{s}_0, a; p_\text{ex}) = y_a),
\end{equation}
where $\mathbf{s}_0$ represents the initialized state with the explicit symptoms, and $\mathbf{y}_\text{im}$ represents the implicit symptoms of the test record $p$, and $D$ is the size of the test set, and $|\mathbf{y}_\text{im}|$ is the number of the implicit symptoms. The better a patient simulator can infer the non-inquired symptoms, the higher the average accuracy will be.

In our experiment, we split the complete dataset (including training samples and test samples) into \textbf{five folds} and calculate the cross-validation IA for each patient simulator. The results and the standard deviation are presented in Tab.~\ref{tab:ia}. According to the results, the vanilla patient simulator is completely non-causal as it fails to infer any symptoms. The generative model learns the correlation without eliminating the collider biases and also demonstrates a quite small IA. In contrast, our PBPS performs much better at inferring correct symptoms.
As shown in Tab.~\ref{tab:ia}, PBPS-PSM performs better than PS and direct estimator, which implies that PSM is less biased than PS and direct estimator. Moreover, we also compare propensity score-vector matching, which uses a vector of propensity scores of all symptom inquiries for matching. Since the score vector is also independent of the inquiry as PBPS, we denote the simulator using propensity score-vector matching as PBPS*. 
From Tab.~\ref{tab:ia}, we can see that without conditioning on the inquiry, PBPS* also performs significantly better than PBPS-PSM, and has a close performance to PBPS. Since the PBPS has a more compact dimension, we use PBPS in the later experiments.

\begin{table}[t]
	\centering
	\setlength{\tabcolsep}{5pt}
	\caption{The quantitative evaluation of the patient simulators by using the SD metric.}
	\begin{tabular}{c|ccccc} \toprule
		Benchmarks & PS~\cite{xu2019end,wei2018task} & GEN~\cite{peng2018deep} & PBPS-PSM & PBPS*  & PBPS  \\ \midrule
		MZ         & 0.085 & 0.081 & 0.174 & 0.391 & 0.425 \\
		DX         & 0.116 & 0.110 & 0.185 & 0.414 & 0.438 \\ \bottomrule
	\end{tabular}
	\label{tab:usersim_tab}
\end{table}

\textbf{Symptom Density.} The symptoms in the collected patient records are sparse and therefore there is a great need for a patient simulator that can infer the unrecorded symptom status \textbf{during training} the interactive MDA agents. Besides the inferring accuracy as measured by IA, we are also interested in the proportion of the answers which are not default answers `not sure', termed as \textit{Symptom Density} (SD). Formally,
\begin{equation}
{\rm SD} = \frac{1}{D\times N_s}\sum_{p \in {\rm train set}}\sum_{a \in [1, ..., N_s]}\mathbb{I}(\mathcal{P}^{\rm train}(\mathbf{s}_T, a; p)\neq 0), \label{equ:sd}
\end{equation}
where $\mathbf{s}_T$ has all explicit and implicit symptoms of the record been observed. The higher SD means the simulator is more likely to answer an inquiry informatively.

SD is calculated among $\mathcal{P}_{\rm PS}^{\rm train}$, $\mathcal{P}_{\rm GEN}^{\rm train}$, $\mathcal{P}_{\rm PBPS-PSM}^{\rm train}$, $\mathcal{P}_{\rm PBPS*}^{\rm train}$ and $\mathcal{P}_{\rm PBPS}^{\rm train}$. As shown in Tab.~\ref{tab:usersim_tab}, our PBPS has obtained the highest score in SD, meaning that our PBPS can generate more informative answers. 

\begin{figure}[t]
	\centering
	\includegraphics[clip=true, trim= 0pt 360pt 600pt 0pt, width=0.8\linewidth]{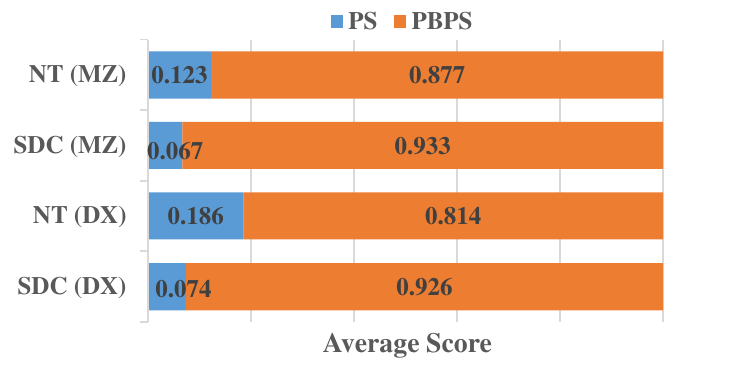}
	\vspace{-10pt}
  \caption{The human evaluation results between PS and PBPS on DX and MZ datasets. NT denotes “naturalness” and SDC stands for “symptom-disease consistency”. The number stands for the average score of preference from the human experts on different metrics and datasets. The higher the score is, the better the patient simulator is.}
	\label{fig:human_evaluation}
\end{figure}
\begin{table*}[t]
	\centering
	\caption{The evaluations of MDA diagnostic success rate, across different RL baselines in \textbf{In.}/\textbf{Out.} settings. }
	\vspace{-5pt}
	\begin{tabular}{c|c|cccc}
		\toprule
		Data           & Setting & DQN~\cite{wei2018task}   & KR-DQN~\cite{xu2019end} & PG-MI-GAN~\cite{xia2020generative} & P2A \\ \midrule
		\multirow{2}{*}{MZ} & \textbf{Out.}    & 0.697\textpm 0.031 & 0.653\textpm 0.016 & 0.721\textpm 0.025  & \textbf{0.784\textpm 0.035} \\
		& \textbf{In.}     & 0.845\textpm 0.026 & 0.755\textpm 0.016  & 0.741\textpm 0.017 & \textbf{0.888\textpm 0.032} \\ \midrule
		\multirow{2}{*}{DX} & \textbf{Out.}    & 0.880\textpm 0.024 & 0.709\textpm 0.011 & 0.676\textpm 0.020 & \textbf{0.928\textpm 0.021} \\
		& \textbf{In.}     & 0.932\textpm 0.018 & 0.777\textpm 0.014  & 0.733\textpm 0.024 & \textbf{0.944\textpm 0.015} \\ \bottomrule
	\end{tabular}
	\vspace{-3mm}
	\label{tab:inout_cmp}
\end{table*}
\begin{table}[!t]
	\centering
	\caption{The MDA diagnostic success rate under \textbf{In.} without inquiry.}
\vspace{-3mm}
	\begin{tabular}{cc|cc}
		\toprule
		\multicolumn{2}{c|}{MZ} & \multicolumn{2}{c}{DX} \\ 
		No intv.   & Intv.   & No intv.    & Intv.    \\ \midrule
		0.935\textpm 0.02     & \textbf{0.972\textpm 0.01}    & 0.971\textpm 0.02  & \textbf{0.981\textpm 0.02} \\ \bottomrule
	\end{tabular}
	\label{tab:no_inq}
\vspace{-3mm}
\end{table}

\textbf{Human evaluation.} 
Besides, we conduct human evaluation between PS and our PBPS to distinguish which simulator is more capable of generating disease-related answers from the angle of the human doctor. We invited six human doctors to interact repeatedly with $\mathcal{P}_{\rm PS}^{\rm all}$ and $\mathcal{P}_{\rm PBPS}^{\rm all}$, and score the \textit{Naturalness} (NT, whose answers are informative like the human patient) and the \textit{Symptom-Disease Consistency} (SDC,  whose answers are more disease-related) for each simulator per evaluation episode. 
More details about the human evaluation are provided in Appendix C.  Since we asked the human participant to choose from two simulators, the scores in Fig.~\ref{fig:human_evaluation} represent the proportions of the preference. As observed in Fig.~\ref{fig:human_evaluation}, the averaging NT and SDC of our PBPS have exceeded the PS sharply, which means in the view of human experts, our PBPS can generate more informative and disease-related answers.

\begin{figure*}[t]
	\centering
	\subfloat{
		\includegraphics[width=0.24\linewidth]{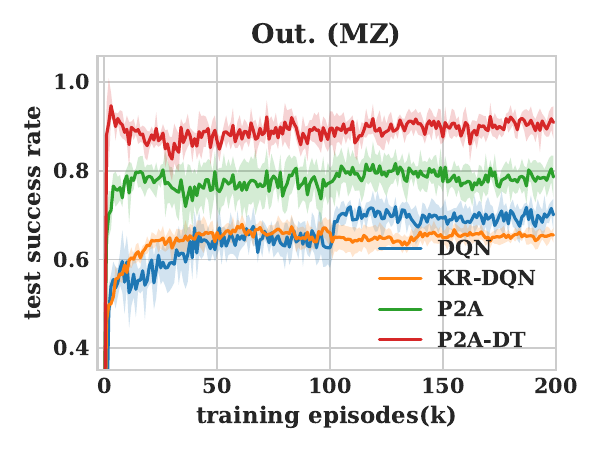}
	}
	\subfloat{
		\includegraphics[width=0.24\linewidth]{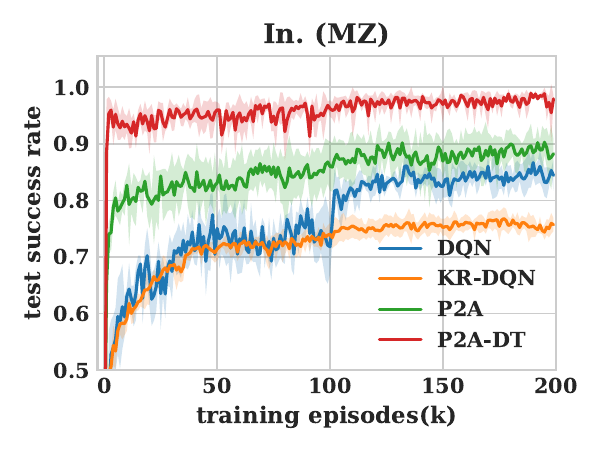}
	}
	\subfloat{
		\includegraphics[width=0.24\linewidth]{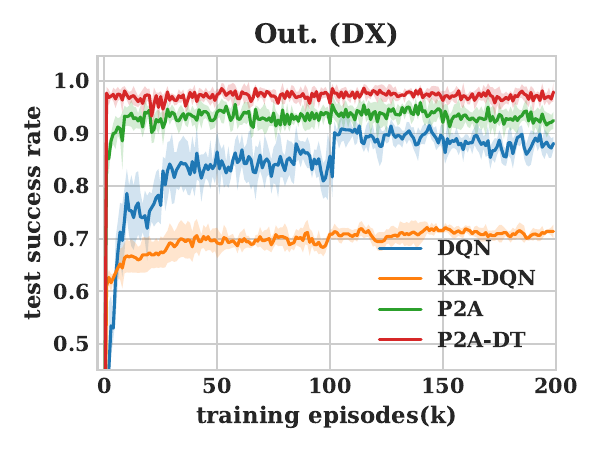}
	}
	\subfloat{
		\includegraphics[width=0.24\linewidth]{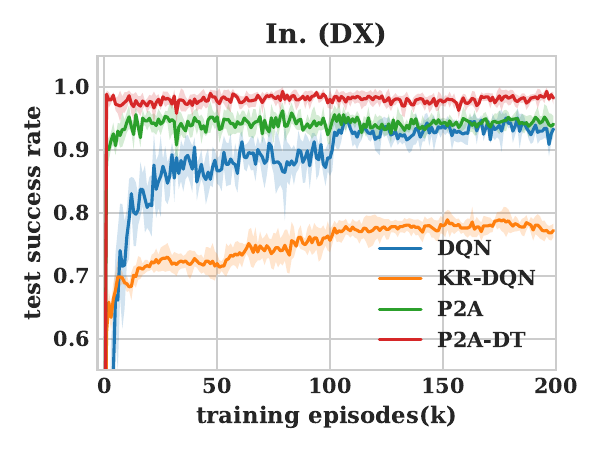}
	}
\vspace{-3mm}
	\caption{The comparison of DQN~\cite{wei2018task}, KR-DQN~\cite{xu2019end}, and P2A on \textbf{MZ} and \textbf{DX} across \textbf{In.}/\textbf{Out.} evaluation settings. The curves denote the mean of the success rate over iterations with deviations.}
\vspace{-3mm}
	\label{fig:inout_cmp}
\end{figure*}
\begin{table*}[!t]
	\centering
	\caption{The evaluations of MDA diagnostic success rate, across different operations in \textbf{In.}/\textbf{Out.} settings. }
\vspace{-3mm}
	\begin{tabular}{c|c|ccc|cc}
		\toprule
		Data           & Setting & No DT (No boot. \& No intv.)  & No bootstrap   & No intervention & P2A & P2A-DT  \\ \midrule
		\multirow{2}{*}{MZ} & \textbf{Out.} & 0.620\textpm 0.03 & \textbf{0.784\textpm 0.05}  & 0.696\textpm 0.03 & \textbf{0.784\textpm 0.04} & 0.91\textpm 0.03 \\
		& \textbf{In.}  & 0.804\textpm 0.02   & 0.834\textpm 0.03 & 0.864\textpm 0.02 & \textbf{0.888\textpm 0.03}  & 0.98\textpm 0.02 \\ \midrule
		\multirow{2}{*}{DX} & \textbf{Out.}  & 0.669\textpm 0.03  & 0.911\textpm 0.02 & 0.840\textpm 0.03 & \textbf{0.928\textpm 0.02} & 0.97\textpm 0.02 \\
		& \textbf{In.}  & 0.788\textpm 0.03   & 0.924\textpm 0.02 & 0.920\textpm 0.03 & \textbf{0.944\textpm 0.02}  & 0.98\textpm 0.01 \\ \bottomrule
	\end{tabular}
\vspace{-3mm}
	\label{tab:bias_cmp}
\end{table*}

\subsection{Evaluation on P2A}
\label{sec:p2a}

As we have answered the first question, we propose empirical studies to demonstrate the superiority of our P2A to answer the other two questions. 

To understand the benefits of the bootstrapping technique and the intervention adopted in our diagnosis agent, we design two test settings, i.e., in-distribution diagnosis (\textbf{In.}) and out-of-distribution diagnosis (\textbf{Out.}), respectively. In the \textbf{In.} setting, the dialogue episodes used for training and testing are all generated by interacting with $\mathcal{P}^{\rm all}_{\rm PBPS}$. This setting aims to evaluate the basic MDA performances without being affected by the distribution bias since the training and testing episodes are from the same distribution. As for the \textbf{Out.} setting, the training dialogue episodes are generated from $\mathcal{P}^{\rm train}_{\rm PBPS}$, then the trained diagnostic agents are tested by interacting with $\mathcal{P}^{\rm all}_{\rm PBPS}$. All the test-simulated patients are invisible during training. Note that all evaluation results are calculated by averaging the results from \textbf{five runs} with different random seeds. The standard deviation is provided in each table and plot~(shadow areas). This setting is to verify the cross-distribution generalization abilities of the diagnostic agent. In this setting, we are interested how the intervention process can help improve transportability. In these experiments, we define a maximum number of inquiries to avoid degenerating the active symptom inquiry into a form-filling manner. And P2A-DT is the diagnostic performance for those cases that can meet DT within the maximum number of inquiries.

\textbf{Accurate and robust diagnosis across \textbf{In.}/\textbf{Out.} settings.}
In Tab.~\ref{tab:inout_cmp}, we compare different baselines by evaluating the mean success rate over the last 20,000 training episodes. Our P2A outperforms the other RL baselines with a clear margin in either \textbf{In.} or \textbf{Out.} setting. Ought to be regarded that, unlike P2A, all other baselines perform very sensitively when the patient simulators are different for training and testing (the \textbf{Out.} setting). In the following, we would like to interpret the results from various aspects and ablate each component of our P2A.

\textbf{Intervention module facilitates the learning of  diagnosers.} In this part, we would like to understand whether the intervention module can help the diagnosers learn to diagnose better. \textit{To investigate the diagnosers solely}, we first train P2A and then take out the bootstrapping diagnosers from P2A for evaluation. To prevent introducing the hint of inquiry behavior in the input states, we fill the non-observed symptoms of the test records using a similar formulation as Equ.~(\ref{equ:sd}) by replacing ``trainset" and ``$\mathcal{P}^\text{train}$" with ``test set" and ``$\mathcal{P}^\text{all}_\text{PBPS}$", respectively. In other words, all unrecorded symptoms in the test set are inquired about at the same time and therefore there is no difference in both the category and order of the inquiries. We then feed these filled states to the diagnosers to measure the diagnostic accuracy, as summarized in Tab.~\ref{tab:no_inq}. From the results, we can observe that the intervention module facilitates the learning of the diagnosers.

\textbf{Intervention module help improve MDA transportability.} To testify whether the intervention can help improve MDA performance in terms of out-of-distribution generalization, we evaluate the diagnostic performance under \text{In.} and \textbf{Out.} settings. The results are presented in the columns of ``No intervention" and ``P2A" (with intervention) of Tab.~\ref{tab:bias_cmp}, which illustrates the averaged effects of  under \textbf{In.} and \textbf{Out.} settings for intervention are 0.024 (=(0.888-0.864+0.944-0.920)/2) and 0.088 (=(0.784-0.696+0.928-0.840)/2). From these results, we observe that the intervention improves significantly under out-of-distribution settings. And we also notice that with intervention, the overall performances in both In. and Out. settings are improved, which might be due to the better learning of the diagnosers with intervention, as we discussed above.

\textbf{Bootstrapping diagnosers help handle the in-distribution diagnosis.} To understand whether the bootstrapping helps improve under \textbf{In.} setting better than \textbf{Out.} setting, we compare our methods with or without either bootstrapping under both settings. The results are presented in columns ``No bootstrapping" and ``P2A" of Tab.~\ref{tab:bias_cmp}., which illustrates that the averaged effects of bootstrapping are 0.037 (=(0.888-0.834+0.944-0.924)/2) and 0.009 (=(0.784-0.784+0.928-0.911)/2) under \textbf{In.} and \textbf{Out.} settings, respectively. From these results, we found that the bootstrapping technique improves the agent more under in-distribution settings. Besides, we noticed under \textbf{Out.} settings of MZ, the performances with and without bootstrapping are nearly the same, which indicates that improvement across distribution is mostly due to the intervention. However, we also notice a case that, under \textbf{In.} on DX, P2A without intervention degrades slightly more than P2A without bootstrapping. The explanation is that the intervention module not only functions in out-of-distribution cases but also helps the diagnosers learn to capture better symptom-disease relationships in general as we analyze above.

\textbf{Decision threshold is the key ingredient of P2A.} In the above discussions, we study the effect of different components on the confidence modeling of the decision threshold. Here, we wonder what if both in-distribution and out-of-distribution confidence are not taken into account. This means the decision threshold is paralyzing since $K\times B =1$ and the standard deviation is 0 in Equ.~(\ref{equ:stati}), which makes the decision threshold always satisfied at the very beginning of the interaction. As shown in column ``No DT (No boot. \& No intv.)" in Tab.~\ref{tab:bias_cmp}, P2A degrades significantly in different settings. From these results, we can see that our decision threshold provides a platform where different modules, i.e., bootstrapping and intervention, can demonstrate their ability in problem-solving.

\textbf{Decision threshold indicates robustness diagnosis.} In Tab.~\ref{tab:bias_cmp}, within the cases that the decision threshold is satisfied (``P2A-DT"), the diagnostic accuracies maintain a very high level, i.e., exceeding 0.9 under different settings. This implies the rationale of using the decision threshold to indicate a robust diagnosis which is quite crucial for the real-world scenario. As also plotted in Fig.\ref{fig:inout_cmp}, P2A-DT appears to possess the abrupt availability of the diagnosis when the decision threshold is satisfied. Fig.\ref{fig:inout_cmp} illustrates the success rate over the iteration of the training episodes required to train the RL agents. P2As (P2A and P2A-DT) achieve faster convergence and higher success rates in upper bounds than all other baselines. Remarkably, those episodes that met the DT achieved a very high success rate even at the very beginning of the training phase~(red curve), meaning that DT only needs a small amount of training data to work reliably (i.e., the abrupt availability of diagnostic knowledge). Such reliable diagnosing performance is significant specifically in the MDA task as the data are expensive to collect.

\begin{figure}[!t]
	\centering
	\subfloat{
		\includegraphics[clip=True, trim=10 0 0 0, width=0.52\linewidth]{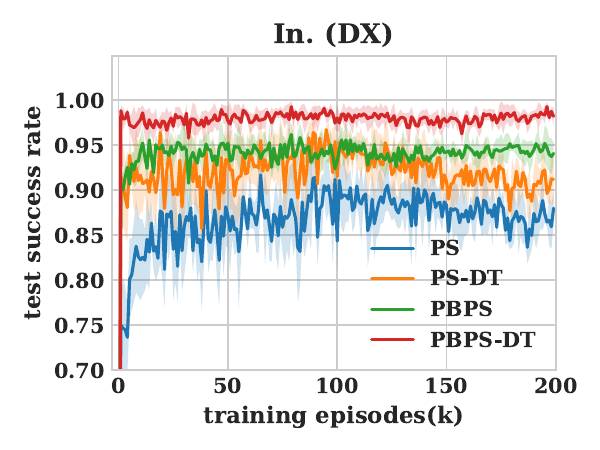}
	}
	\subfloat{
		\hspace{-15pt}
		\includegraphics[clip=True, trim=22 0 0 0, width=0.50\linewidth]{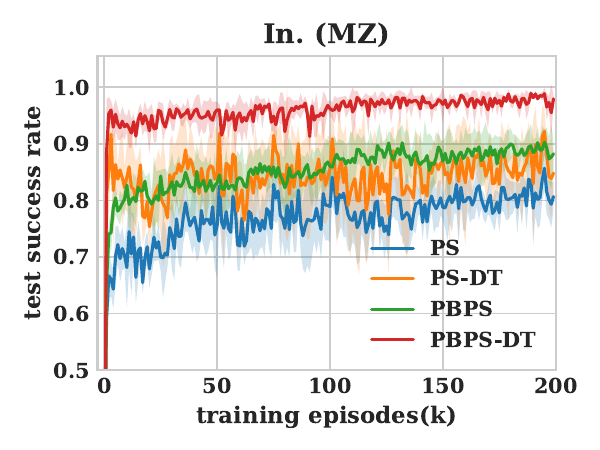}
	}
	\vspace{-3.5mm}
	\caption{Performance of P2A with PS and P2A with our PBPS under \textbf{In.} setting with DX and MZ datasets.}
	\label{fig:dx_in}
\end{figure}

\textbf{PBPS is beneficial to P2A with diagnosis accuracy and uncertainty modeling.} To testify whether our PBPS is beneficial to P2A from both diagnosis performance and the accuracy of uncertainty modeling, we have also evaluated P2A with PS and our PBPS under the in-distribution setting, as shown in Fig.~\ref{fig:dx_in}. `PS' denotes P2A trained with PS. `PS-DT', `PBPS' as well as `PBPS-DT' follow the same denotation rules. From the results, we observe that without our PBPS, the performance of P2A dropped sharply. Especially, under \textbf{In.} setting, `PBPS-DT' is able to achieve near-perfect results, meaning that with responses from our PBPS, the uncertainty modeling of P2A can almost capture the actual symptom-disease relation. Besides, the improvement of accuracy with `DT' is much more obvious for `PBPS', which implies that the agent trained with our PBPS gains a sharp improvement performance with better uncertainty estimation.

\begin{table}[t]
\centering
\caption{The performance of P2A with different scales of $\sigma$ of the decision threshold.}
\vspace{-3mm}
\label{tab:scale}
\begin{tabular}{c|c|cccc}
\toprule
Data                & Setting & $1\sigma$ & $3\sigma$ & $6\sigma$ & $12\sigma$ \\ \midrule
\multirow{2}{*}{MZ} & \textbf{Out.}     & 0.69\textpm 0.03  & 0.77\textpm 0.04  & \textbf{0.78\textpm 0.04}  &   0.75\textpm 0.08  \\
                    & \textbf{In.}      & 0.83\textpm 0.03  & 0.88\textpm 0.03  &  \textbf{0.89\textpm 0.03} &   0.84\textpm 0.07  \\ \midrule
\multirow{2}{*}{DX} & \textbf{Out.}     & 0.69\textpm 0.03  &  \textbf{0.93\textpm 0.02} & \textbf{0.93\textpm 0.02}  &  0.90\textpm 0.08  \\
                    & \textbf{In.}      & 0.86\textpm 0.02  & \textbf{0.94\textpm 0.02}  & \textbf{0.94\textpm 0.02}  &    0.91\textpm 0.08 \\ \bottomrule
\end{tabular}
\vspace{-3mm}
\end{table}

\textbf{Different scales of $\sigma$ for the decision threshold.} In this part, we study the selection of different scales of $\sigma$ for the decision threshold. The experimental results are listed in Tab.~\ref{tab:scale}. From the results, we observe that, besides some extreme cases (i.e., 1$\sigma$ or 12$\sigma$), P2A doesn’t appear to be sensitive to the scales of $\sigma$. The potential reason is that the diseases are distinguishable as long as the necessary symptom information has been observed. In this sense, if the scale of $\sigma$ is large enough for driving P2A to collect the necessary information for diagnosing, P2A would perform about the same. In this sense, if the scale is too small, e.g. 1$\sigma$, P2A will be unable to collect enough information for diagnosing. However, we also observe that when the scale is too large, e.g., 12$\sigma$, the decision threshold becomes difficult to meet which might hamper the learning of inquiry policy.

\textbf{Cross-dataset generalization.} Although different datasets are collected from different researchers~\cite{wei2018task,xu2019end}, we found that there are two shared diseases of the two datasets, i.e., i.e., upper respiratory tract infection, and infantile diarrhea. To this end, some might be interested in how different methods perform when training in DX and testing in MZ in these two shared diseases, and vice versus. To align different types of symptoms in different datasets, we make some modifications that match symptoms in the two datasets according to the semantic similarity (the L2-distance of embeddings extracted by BERT-tiny~\cite{devlin2018bert}). After matching, we simply rename the symptoms in one dataset with their most similar symptoms in the other. We use MZ$\rightarrow$DX to denote the dataset which is MZ originally and is revised to DX by renaming, and so is DX$\rightarrow$MZ. The different MDA agents are trained in the in-distribution setting using the original datasets. As for cross-dataset evaluation, all data from MZ$\rightarrow$DX and DX$\rightarrow$MZ are used. The results are shown in Tab.~\ref{tab:me}. From the results, It can be seen that our P2A can still obtain the best performance in cross-dataset generalization.

\begin{table}[t]
\caption{Cross-dataset generalization of P2As trained on MZ/DX and evaluated on DX$\rightarrow$MZ/MZ$\rightarrow$DX.}
	\vspace{-3mm}
\setlength{\tabcolsep}{1mm}
\label{tab:me}
\centering
\begin{tabular}{c|cccc} \toprule
Ori. $\rightarrow$ revise   & DQN~\cite{wei2018task}   & KR-DQN~\cite{xu2019end} & PG-MI-GAN~\cite{xia2020generative} & P2A \\ \midrule
MZ$\rightarrow$ DX &  0.79\textpm 0.05   &   0.66\textpm 0.04     &   0.71\textpm 0.05        &  \textbf{0.87\textpm 0.02}    \\
DX$\rightarrow$ MZ &   0.63\textpm 0.04  &   0.59\textpm 0.03     &   0.65\textpm 0.04        &   \textbf{0.76\textpm 0.04}  \\ \bottomrule
\end{tabular}
\vspace{-3mm}
\end{table}

\section{Conclusion}
This paper presents a complete framework for the MDA task, CA-MDA, including a simulator PBPS that tackles the problem of counterfactual symptom inquiry by PLM, and an MDA agent P2A that additionally eliminates the distributional bias via intervention and models the confidence to drive the symptom inquiry. Through mining the passive observational data in such a collaborative representation manner, to the best of our knowledge, this paper is the first study to propose a simple yet concise paradigm to address sequential discrimination tasks with only passive observational data. Experimental results demonstrate that PBPS can generate more informative and disease-related answers. Moreover, P2A is more accurate and robust across distributions. Our introduced Decision Threshold provides a reliable stop mechanism for MDA agents. In the future, we will study how to extend our framework to broader ranges of sequential discriminative decision-making problems with passive observational data.

\section*{Acknowledgements}
This work was partly supported by the National Key R\&D Program of China under Grant No. 2021ZD0111601, National Natural Science Foundation of China (NSFC) under Grant No.61836012 and No.62276283, Guangdong Basic and Applied Basic Research Foundation (No.2023A1515011374 and No.2023A1515012985), Basic and Applied Basic Research Special Projects under Grant No. SL2022A04J01685.

\ifCLASSOPTIONcaptionsoff
  \newpage
\fi

\appendices

\begin{figure*}[t]
	\centering
	\footnotesize
	\includegraphics[clip=True, trim=0 260pt 0 0, width=\linewidth]{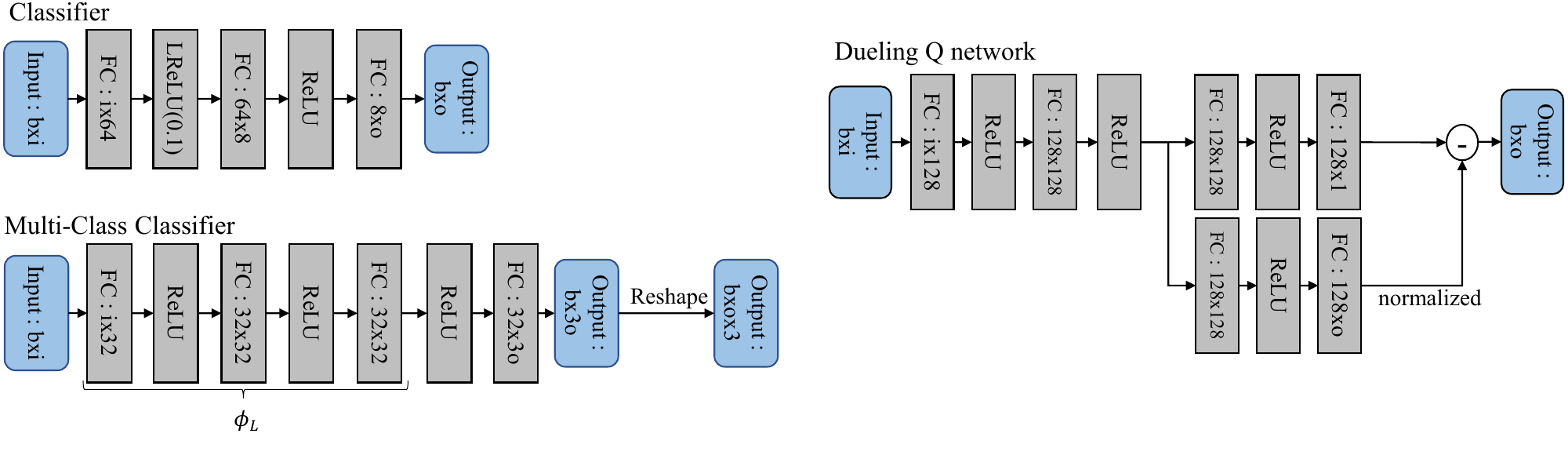}
	\hspace{-10pt}
	\caption{\footnotesize Two types of network structures used throughout the whole work, i.e. "Classifier"~(upper) and "Multi-Class Classifier"~(lower). "FC" is the abbreviated form of a fully connected layer, and "LReLU~(0.1)" is the abbreviated form of leaky ReLU activation with a negative slope of 0.1. $b$ denotes the batch size of the input, $i$ denotes the input dimension and $o$ stands for the output dimension.}
	\label{fig:network}
	\vspace{-5pt}
\end{figure*}

\section{Propensity-based Patient simulator~(PBPS)}

\subsection{Propensity Score Matching~(PSM)}
In real life, we can only observe the outcome of the actual action, which is named as the factual outcome. However, sometimes we might wonder what the outcome should have been if the other action had been taken, that is the counterfactual outcome. The potential outcome can be either factual or counterfactual. According to the potential outcome framework~\cite{rubin1974estimating,neyman1923applications}, the counterfactual outcome is inferable if the three assumptions are met: the stable unit treatment value assumption (SUTVA), consistency and ignorability~(unconfoundedness). Ordinarily, SUTVA and consistency are assumed to be satisfied. With ignorability, we assume that all the confounding variables are observed and reliably measured by a set of features $X^{(u)}$ for each instance $u$. $X^{(u)}$ denotes a set of confounding variables, namely a subset of features that describes the instance $u$ and causally influences the values of both the treatment $A^{(u)}$ and the potential outcome $Y_a^{(u)}$. Ignorability means that the values of the potential outcomes $Y_a^{(u)}$ are independent of the factual action $A^{(u)}$, given $X^{(u)}$. Mathematically, ignorability can be formulated as:
\begin{equation}
Y_a^{(u)} \indep A^{(u)} | X^{(u)}.
\end{equation}
From the notation, we can see that this is an assumption defined at the individual level. With ignorability satisfied, we can estimate the counterfactual outcome of $u$ through the factual outcomes of other instances $u^\prime$ with the same covariates  $Y_a^{(u)}=Y_a^{(u^\prime)}|X^{(u)}=X^{(u^\prime)}$, which is the essence of \emph{the matching methods}~\cite{stuart2010matching}. 

However, we need to be careful when there exists a group that only contains instances with only one type of action. We cannot estimate the counterfactual outcome in this group. This issue is referred to as \emph{the lack of overlap}. To overcome this issue, the most widely adopted matching as weighting methods specify the function $e(X^{(u)})$ in estimating \emph{propensity score} $P(A^{(u)}|X^{(u)})$~\cite{austin2011introduction,rosenbaum1983central}. We need to make sure whether the ignorability still holds with $e(X^{(u)})$. In other words, given $e(X^{(u)})$, whether the $Y_a^{(u)} \indep A^{(u)} | e(X^{(u)})$ still holds. We show that $Y_a^{(u)} \indep A^{(u)} | e(X^{(u)})$ is satisfied in the following. Here, for briefness, ($Y, X, A$) denotes $(Y_a^{(u)}, X^{(u)}, A^{(u)})$.
\begin{equation}
\begin{aligned}
P(A | e(X)) & = \mathbb{E}[A|e(X)] = \mathbb{E}[\mathbb{E}[A|X, e(X)]|X] \\
& = \mathbb{E}[\mathbb{E}[A|X]|X] = \mathbb{E}[e(X)|X] =e(X)
\end{aligned}
\label{equ:2}
\end{equation}
\begin{equation}
\begin{aligned}
P(Y, A | &e(X)) = P(Y|e(X))P(A|Y, e(X)) \\
&= P(Y|e(X)) \mathbb{E}[P(A|Y, e(X), X)|X] \\
&= P(Y|e(X))\mathbb{E}[P(A|Y, X)|X] \\
&= P(Y|e(X)) \mathbb{E}[P(A|X)|X] \quad \text{by ignorability} \\
&= P(Y|e(X)) \mathbb{E}[e(X)|X] \\
&= P(Y|e(X))e(X) \\
&= P(Y|e(X))P(A|e(X)) \quad \text{according to (\ref{equ:2})}
\end{aligned}
\label{equ:4}
\end{equation}

From (\ref{equ:4}), we know that $Y_a^{(u)} \indep A^{(u)} | e(X^{(u)})$. It means that we can estimate the counterfactual outcome of $u$ by matching instances with the same propensity score $e(X^{(u)})$.

\subsection{Details of PBPS}

The multi-classifier adopted for training the propensity score $e_{\phi_{\rm S}}\big(f_{\phi_{\rm L}}(\mathbf{y}, d), a\big)$ with parameter $\phi_{\rm S}$ and $\phi_{\rm L}$ is shown in Fig.~\ref{fig:network}, whose intermediate result $f_{\phi_{\rm L}}(\mathbf{y}, d)$ from the second to the last fully connected (FC) layer is the propensity latent features. The input dimension \textit{i} is set as $N_s+N_d$ and the output dimension \textit{o} is set as $N_s$, where $N_d$ is the number of diseases, and $N_s$ is the number of symptoms. 
The propensity of a record $q$ according to $p$ is $f_{\phi_{\rm L}}(\mathbf{y}^{(q)}\odot\mathbf{m}^{(p)}, d^{(q)})$, 
where $\mathbf{m}^{(p)} = [\mathbb{I}(y_a^{(p)} \neq 0), \forall a \in [1, ..., N_s]]$.

\section{Progressive Assurance Agent~(P2A)}

\subsection{State representation}
During training our P2A, the dialogue state has the value mapping \{“no”:-1, “not sure”:0, “yes”:1\}. The reason why we use this value mapping is two-fold: 1) the values (-1, 0, 1) well-align the semantic relationship among “no”, “not sure” and “yes”. Specifically, “no” has the opposite representation of “yes”, and “not sure” is in between; 2) leveraging the semantic relationship among different statuses can also share information across data. Especially when the data is sparse, some symptom statuses might rarely or never appear during training, using one-hot encoding will cause their corresponding parameters to rarely or never be updated during training, which might cause unexpected results during deployment. And to indicate which symptoms are visited, we also use an extra binary vector $\mathbf{m}$ in which the visited symptoms have value 1. And in our P2A, the binary vector is used in the inquiry policy to avoid repeat inquiries.

\subsection{Details of P2A}

\subsubsection{Intervener} 
\begin{figure}[t]
	\centering
	\includegraphics[clip=true, trim= 0pt 930pt 400pt 0pt, width=\linewidth]{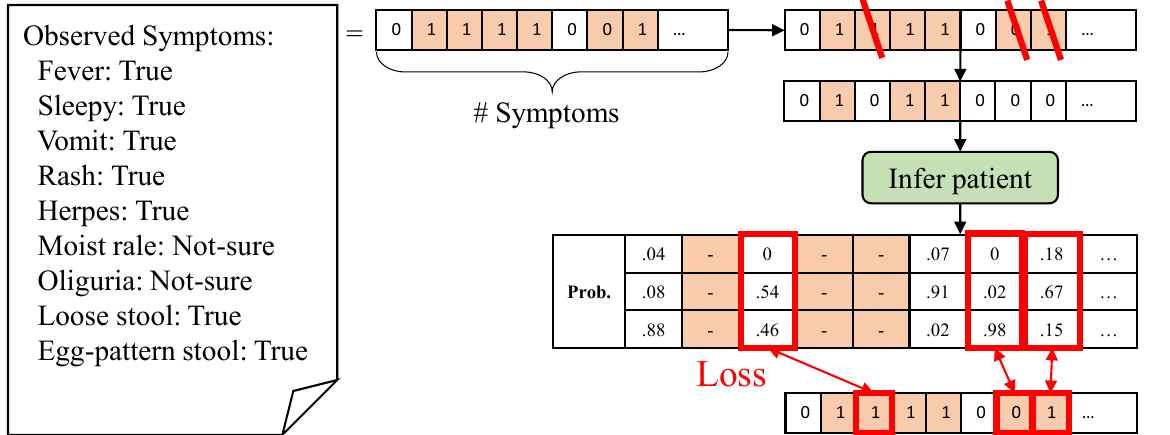}
	\caption{Training strategy of the intervener. The inquired information is masked off and the intervener is trained to recover them back.}
	\label{fig:int_train}
\end{figure}

Similar to the PBPS, we adopt the multi-classifier network in Fig.~\ref{fig:network} for the intervener. This intervener is aimed at predicting the final state $\hat{\mathbf{s}}_T$ according to the current visited symptoms $\mathbf{s}_t^v$. Therefore, we model it as a generative problem. $\phi_{\rm G}$ is the parameter of the generator $f_{\phi_{\rm G}}(\cdot)$. The input dimension \textit{i} and the output dimension \textit{o} are both set as $N_s$. During the training phase, At the end of each diagnosis, the final state and the ground-truth disease are stored in a final state buffer. To train the intervention module, we sample the final states from that buffer. Before input to the diagnoser, we randomly mask off several observed symptoms of the states. We only calculate the cross-entropy loss for the masked symptoms since we only have labels for these symptoms. An intuitive demonstration of the training process is as Fig.~\ref{fig:int_train}.

\subsubsection{Bootstrapping diagnosers} The network structure of the bootstrapping diagnoser is the classifier structure in Fig.~\ref{fig:network}. The input dimension is set as $N_s$, and the output dimension is set as $N_d$. Besides the final states stored in the final state buffer, we also sample states from the interveners to extend the training data. Both final states and intervened states are used to train the diagnosers.

\subsubsection{Policy network} The Q-learning policy network we adopted is the Dueling-DQN~\cite{wang2015dueling} in Fig.~\ref{fig:network} for stabilizing the training procedure, which has two heads with one for state value $V$ and the other one for action advantage value $A$. Therefore, $Q(\mathbf{s}_t, \boldsymbol{\mu}_t, a_t) = V(\mathbf{s}_t, \boldsymbol{\mu}_t)+A(\mathbf{s}_t, \boldsymbol{\mu}_t, a_t)$, where $\boldsymbol{\mu}_t$ is the expectation of disease probability calculated from the output of the bootstrapping diagnosers. In order to avoid repeating choosing symptoms that have been visited, we subtract 1000000 for each symptom that has been inquired from the $Q(\mathbf{s}_t, \boldsymbol{\mu}_t, a_t)$ to decrease the q values of these symptoms. The input contains three vectors, i.e., the state $\mathbf{s}_t$, the inquired history vector (mentioned in Sect. B.1), and the expectation of diseases' probabilities $\boldsymbol{\mu}$. Therefore, the input dimension \textit{i} is set as $2\times N_s+N_d$ and the output dimension \textit{o} is set as $N_s$.

\section{Experiment Details}

\subsection{Benchmark Examples}

\begin{figure}[t]
	\centering
	\includegraphics[clip=true, trim=0pt 80pt 680pt 0pt, width=0.9\linewidth]{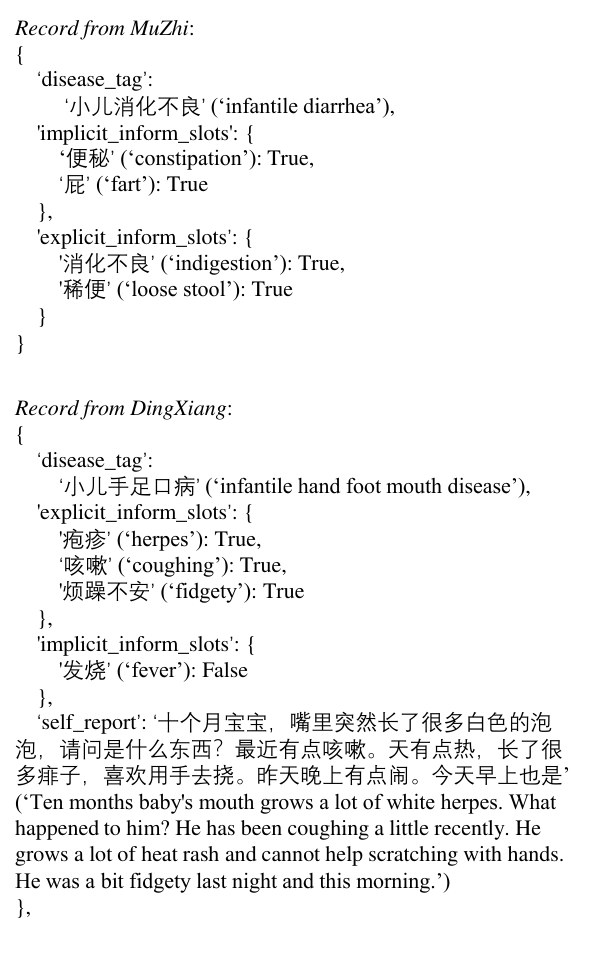}
	\vspace{-10pt}
	\caption{The record examples from benchmarks MuZhi~(MZ)~\cite{wei2018task} and DingXiang~(DX)~\cite{xu2019end}. The upper record is sampled from MZ, which is in a structural and clean format. The bottom record is sampled from DX, different from MZ, whose original self-report is preserved. Both benchmarks are in Chinese and we present them in English in this paper.}
	\vspace{-5pt}
	\label{fig:record}
\end{figure}

In this part, we briefly describe the data format of two open benchmarks, MZ~(MuZhi)~\cite{wei2018task} and DX~(DingXiang)~\cite{xu2019end}. The diagnosis record from MZ is clean and structural in which original dialogue sentences are not preserved. Different from {MZ and DX that} preserve the original sentences of the self-report. In both datasets, the record contains the ground-truth disease ``\verb|disease_tag|", the symptom information extracted from the original self-report ``\verb|explicit_inform_slots|", and also the symptom information mentioned during the original dialogue ``\verb|implicit_inform_slots|". Two examples from these datasets are demonstrated in Fig.~\ref{fig:record}. Notice that, these two datasets are in Chinese originally and we translate them into English to demonstrate in this paper.

\begin{figure}[t]
	\centering
	\subfloat[]{
		\includegraphics[width=\linewidth]{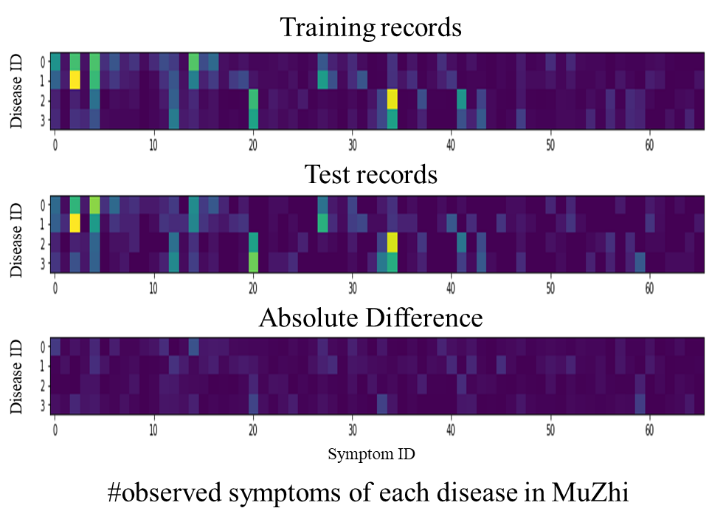}
	} \\
	\subfloat[]{
		\includegraphics[width=\linewidth]{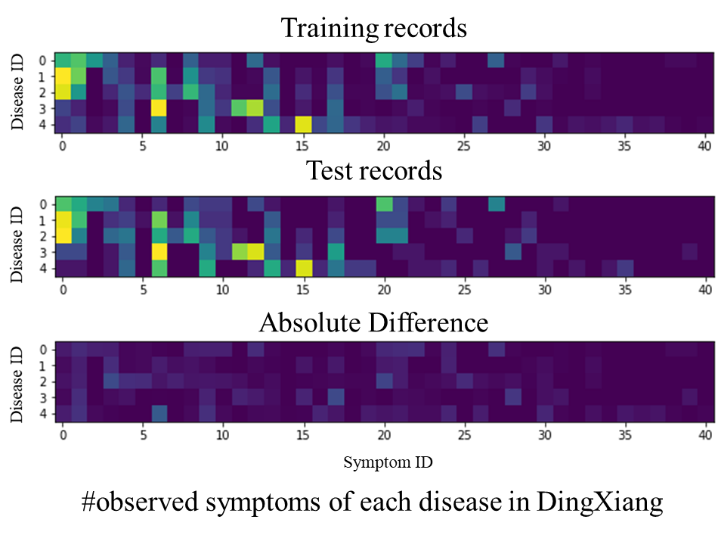}
	}
	\caption{The illustration of the distribution of recorded symptoms. In these plots, the brighter the color means the more frequent the symptoms are observed. For each dataset, we plot the symptom distributions of the train and test splits. Besides, we also show the absolute difference between the train and the test splits. The upper three plots are for the MuZhi data set and the lower three are for DingXiang.}
 \label{fig:dist}
\end{figure}

In the MZ dataset, there are 4 diseases (infantile bronchitis, upper respiratory tract infection, infantile diarrhea, and infantile dyspepsia) and 66 symptoms. The numbers of training records for the diseases are 130, 166, 155, and 117. The numbers of test records are 30, 34, 45, and 33. As for the DX dataset, there are 5 diseases (allergic rhinitis, upper respiratory tract infection, infantile diarrhea, infantile hand-foot-and-mouth disease, and pneumonia) and 41 symptoms. The numbers of training records for the diseases are 82, 98, 80, 81, and 82. And the numbers of test records are 20, 24, 20, 20, and 20. To show the distribution of symptoms for each disease, we calculate the number of observed symptoms for each disease and normalized them into a range [0, 1], as plotted in the upper part of Fig.~\ref{fig:dist}. To show the distribution gap between training and test datasets, we also calculate the absolute difference between the two normalized results.

From Fig.~\ref{fig:dist}, we can see that the numbers of records for different diseases in the training/test set are relatively balanced. And we can observe that the ratios of observed symptoms of each disease are similar between training and test sets. However, there still have noticeable gaps between the symptom ratios between training and test sets.

\begin{table*}[t]
	\centering
	\caption{The statistical characteristics of the human evaluation of different participants.}
	\begin{tabular}{c|cccccc}
		\hline
		Participant ID & 1             & 2          & 3   & 4           & 5   & 6           \\
		\hline
		NT~(MZ)        & 0.82\textpm 0.38   & 1\textpm 0        & 1\textpm 0 & 0.90\textpm 0.31 & 1\textpm 0 & 0.62\textpm 0.49 \\
		SDC~(MZ)       & 0.94\textpm 0.24   & 1\textpm 0        & 1\textpm 0 & 0.90\textpm 0.31 & 1\textpm 0 & 0.81\textpm 0.39 \\
		NT~(DX)        & 0.52\textpm 0.50   & 1\textpm 0        & 1\textpm 0 & 0.68\textpm 0.47 & 1\textpm 0 & 0.69\textpm 0.46 \\
		SDC~(DX)       & 0.89\textpm 0.31   & 0.96\textpm 0.20 & 1\textpm 0 & 0.86\textpm 0.35 & 1\textpm 0 & 0.85\textpm 0.36 \\
		\hline
	\end{tabular}
	\label{tab:human}
\end{table*}

\begin{table}[!t]
	\centering
	\caption{The statistical hypothesis testing~(P-value) of different hypotheses.}
	\begin{tabular}{ccccc}
			\hline
			Hypothesis & NT~(MZ)    & SDC~(MZ)    & NT~(DX)    & SDC~(DX)   \\
			\hline
			P-value    & 5.4e-21 & 1.2e-33 & 9.0e-20 & 1.6e-47 \\
			\hline
	\end{tabular}
	\label{tab:pvalue}
\end{table}

\subsection{Training Details}
For training the propensity score/latent feature estimator of PBPS, the learning rate is 0.01 initially and is decreased to the tenth of it for every 10000 iterations. The total training iterations is 40000 with the Adam optimizer. The batch size b is 128.

For training the P2A, the learning rate of the policy network is 0.001 and is decreased to the tenth of it after 100,000 training episodes while the learning rate for the intervener and the bootstrapping diagnosers is fixed at 0.001. All of the parameters are updated by the Adam optimizer. The maximum number of training episodes is 200,000. The max dialogue rounds $T$ is 1/3*N. The constant reward -0.1 is given to the agent for each round to encourage shorter dialogues. The $\epsilon$ of $\epsilon$-greedy strategy is set to 0.1 for efficient action space exploration, and the discount factor $\gamma$ in the Bellman equation is 0.95. The size of buffer $\mathcal{D}_{Q}$ is 50,000 and the size of buffer $\mathcal{D}_{C}$ is 1280. The batch size is 32, and the Polyak factor is 0.99. The number of samples of the intervener is set as 50, and the number of the bootstrapping models is 10. We update the target policy every ten rounds and update the model every round after 6,000 random start rounds.

\subsection{Baselines details}
The RL baselines DQN~\cite{wei2018task} and KR-DQN~\cite{xu2019end} are trained with the open-source code from KR-DQN~\cite{xu2019end} with default training settings. 
Moreover, the batch size for training is set as 32, and the learning rate is 0.001, which will decrease to the tenth of it after every 40000 iterations with the Adam optimizer. The total training iterations is set as 100000. 

\subsection{Evaluation for Patient Simulator}

\subsubsection{Human evaluation details}
Although we have designed several quantitative metrics to evaluate the patient simulators, the quantitative metrics still have limitations to reflect “which one is more disease-related” from the perspective of human experts. In order to measure the qualitative performance of the patient simulators, i.e. PS and PBPS, we propose `Naturalness'~(NT) and `Symptom-Disease Consistency'~(SDC) metrics. NT is to score how natural the simulator is (we ask the human experts: "Which one is more natural?") and SDC is to score whether the patient simulator can generate disease-related responses (we ask the human experts: "Which one is more disease-related?"). Although these questions seem vague and general, they are exactly the questions we are interested in. And any additional remarks/conditions will cause the answers biased from the question. The reason that we use NT instead of informativity~(corresponding to `informative') is that informativity can be well quantified by the `Symptom Density' proposed in the quantitative evaluation. Realizing that the ultimate goal of the quantitative and qualitative evaluations is to find out which patient simulator is more like a human patient. we decide to employ the more essential yet more subjective index, i.e., `Naturalness'.

In the human evaluation, six physicians were invited, who are all highly trained and credentialed. They have all graduated from the Sun Yat-sen University School of Medicine. Among them, two hold Ph.D. degrees, indicating their advanced training and research expertise in their respective medical fields. The other four have earned Master's degrees, underlining their comprehensive understanding and considerable proficiency in medicine. All six physicians possess the qualification certificate of clinical medicine practicing physician. 
The experts were required to decide which simulator is more natural (score 1) and which simulator is more disease-related (score 1) per episode with the two patient simulators. Each participant was asked to interact with the two patient simulators with 30 episodes~(5 interactions each episode) with random anchor records from each dataset. We adopted the simple template-based natural language generator for both the patient simulators to respond to the doctor's inquiry simultaneously. As for the doctors, we offered them the list of symptoms of each dataset. For each evaluation episode, one random anchor record was chosen for both the patient simulators, and the doctors were provided with the self-report and the corresponding ground truth disease of the anchor record. Then the doctor would initiate the inquiries about symptoms which would be answered by both patient simulators without knowing which answer is from which patient simulator. From the above description, we know that 1) the human participant proactively selected symptoms to inquire about without knowing the occurrence frequency of symptoms; 2) all symptoms were not presented according to their occurrence frequency; 3) the human participants are responsible to grade which simulator can render disease-related answer. Therefore, the evaluation process is independent of the occurrence frequency of symptoms. And only human experts are responsible for deciding which simulator can render disease-related answers.

\begin{figure*}[t]
	\centering
	\subfloat[]{
		\includegraphics[width=0.24\linewidth]{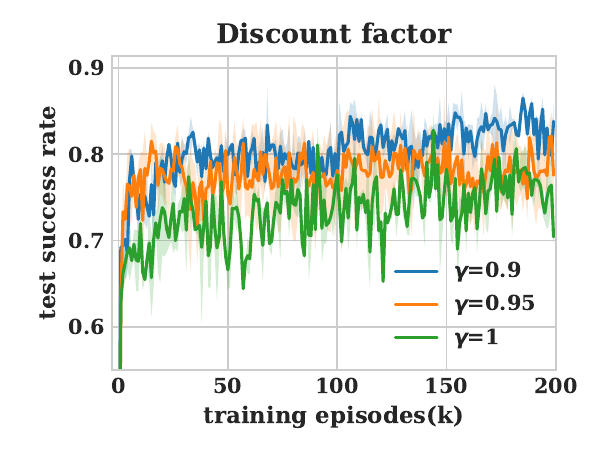}
	}
	\subfloat[]{
		\includegraphics[width=0.24\linewidth]{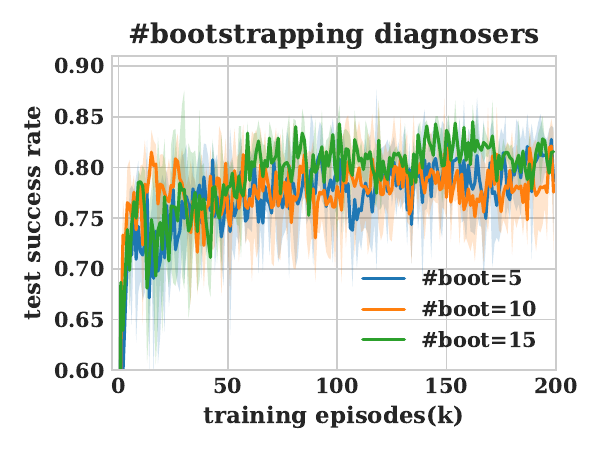}
	}
	\subfloat[]{
		\includegraphics[width=0.24\linewidth]{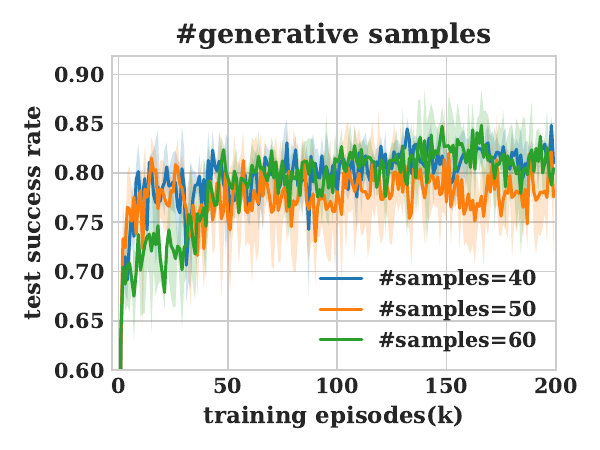}
	}
	\subfloat[]{
		\includegraphics[width=0.24\linewidth]{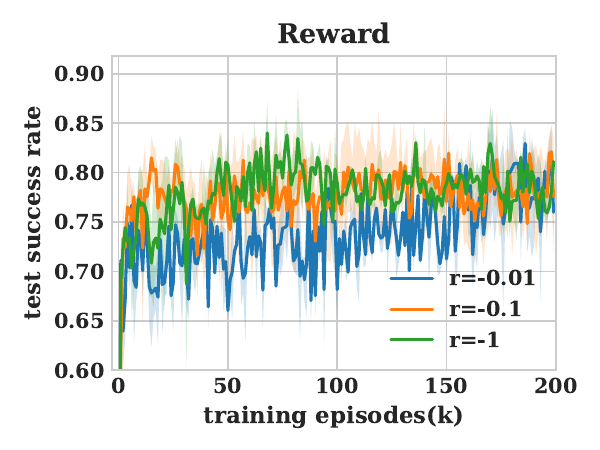}
	}
	\caption{The success-rate curves of our P2A across different hyperparameters on MZ under Out. setting. From left to right, the figures are for different discount factors, the number of bootstrapping diagnosers, the number of samples of bootstrapping diagnosers, and rewards. The curves of the default hyperparameters are orange.}
	\label{fig:hyperp}
	\vspace{-10pt}
\end{figure*}

After interviewing, we counted the proportion of NT and SDC for each patient simulator by dividing the total number of samples, i.e. 6*30~(6 doctors and 30 samples for each dataset and each patient simulator). 
In order to provide more comprehensive statistical results of the human evaluation, we also provided the mean and standard deviation of scores from different human participants in Tab.~\ref{tab:human} and the P-value~\cite{johnson2000probability} of each human evaluation result as shown in Tab.~\ref{tab:pvalue}. 1 denotes our approach better, 0 otherwise. In Tab.~\ref{tab:human}, a value greater than 0.5 means our simulator is better (0.5 means tie statistically). As can be seen from the table, for each participant and each measurement, our proposed method always scores higher statistically. Notice that since the score is a binary value from {0, 1}, the standard deviation becomes larger when the expectation approaches 0.5. In Tab.~\ref{tab:pvalue}, the null hypothesis of each index is that two patients are assumed to be even, and the expectation value is 0.5. The p-value is one-tailed since we only need to consider how significant the claim is that our method is better than the original. The significance level is set as 0.05 empirically~\cite{johnson2000probability}. From the results in the table, we can observe that all p-values are pretty small, which means our claims are always significant across different benchmarks and measures.

\begin{figure*}[t]
	\centering
	\subfloat[]{
		\includegraphics[width=0.24\linewidth]{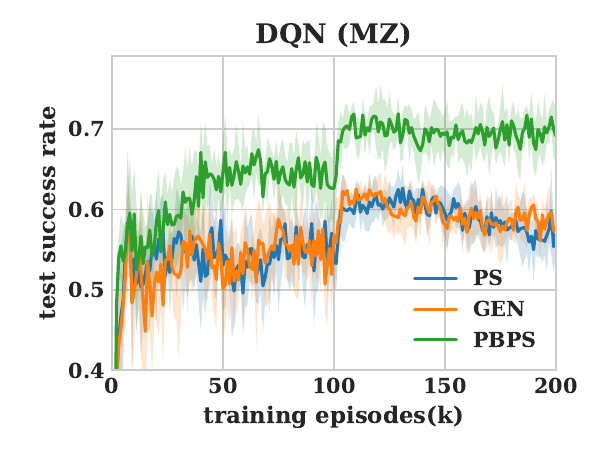}
	}
	\subfloat[]{
		\includegraphics[width=0.24\linewidth]{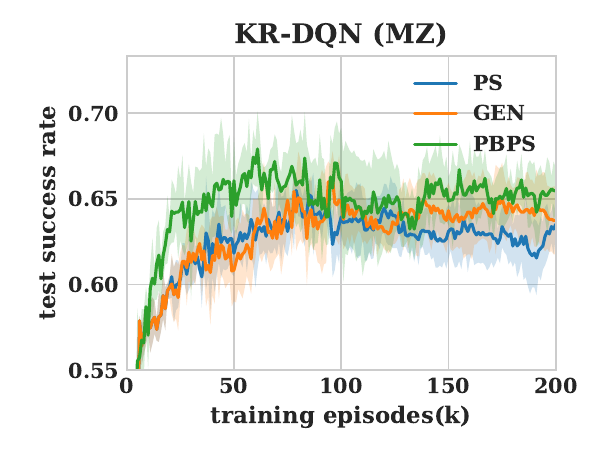}
	}
	\subfloat[]{
		\includegraphics[width=0.24\linewidth]{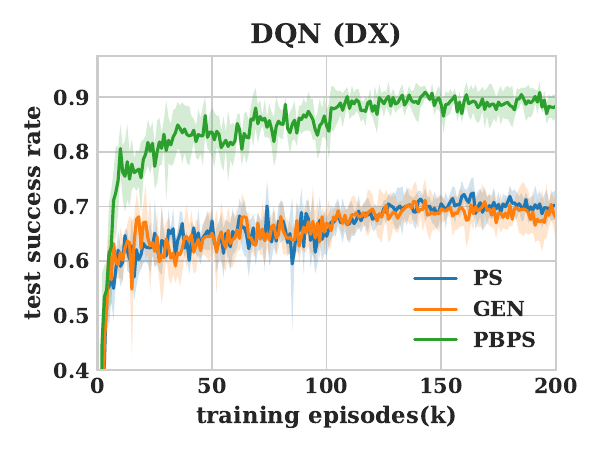}
	}
	\subfloat[]{
		\includegraphics[width=0.24\linewidth]{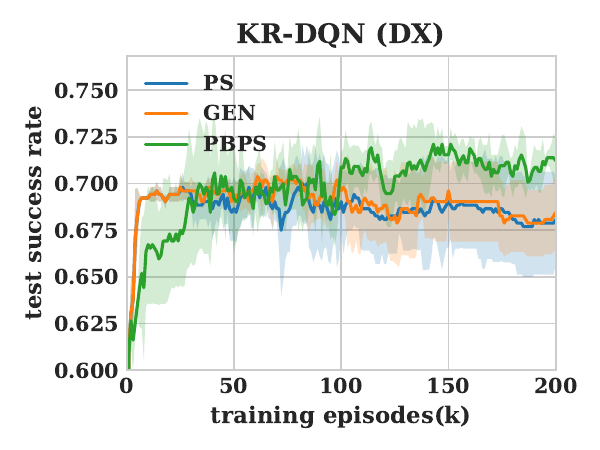}
	}
	\caption{The success-rate curves of the original patient simulator (PS)~\cite{xu2019end,wei2018task}, generative world model (GEN)~\cite{peng2018deep}, and our probabilistic-symptom patient simulator (PBPS) based on DQN~\cite{wei2018task} and KR-DQN~\cite{xu2019end} trained in the benchmarks MZ and DX, respectively.}
	\label{fig:usersim_cmp}
\end{figure*}

\subsection{More Experimental Results}

\textbf{Selection of hyperparameters.} In this part, we conduct extra ablation studies of our proposed method to understand the sensitivity of different hyperparameters. Especially, despite those basic hyperparameters such as learning rate and decay frequency that are used in all baselines, there are several important hyperparameters that are special in our methods. They are discount factor $\gamma$ adopted in Bellman backup, the number of bootstrapping diagnosers for estimate uncertainty, the generative sample number of the intervener, and the reward signal designed to drive the learning process. For the exposition, we evaluate our method on the MZ dataset under the out-of-distribution setting, as it should be enough to demonstrate the robustness of our method. The performance of the different values of these hyperparameters is plotted in Fig.~\ref{fig:hyperp}. From this figure, we can observe that our proposed method is not sensitive to the selection of hyperparameters and all the results are superior to the baseline methods.

\textbf{Different similarity metric for propensity latent matching.} Besides L2-norm distance, we also conduct experiments with L1-norm and cosine similarity to assess the similarity of two latent features. The ablation results are shown in Tab.~\ref{tab:sm}. And we observed that the results between different measurements were close, and L2-norm obtain the best results.

\begin{table}[t]
\centering
\caption{Different IA results of PBPS with different similarity measurements.}
\begin{tabular}{c|ccc}
\hline
dataset & L2-norm & L1-norm & cosine similarity  \\ \hline
MZ      &  \textbf{0.62\textpm 0.02}   &  0.60\textpm 0.02     &  0.59\textpm 0.02                  \\
DX      & \textbf{0.66\textpm 0.01}   &  0.64\textpm 0.02       &   0.64\textpm 0.02                 \\ \hline
\end{tabular}
\label{tab:sm}
\end{table}

\begin{figure*}[t]
	\centering
	\includegraphics[clip=true, trim= 0pt 120pt 8pt 0pt, width=0.7\linewidth]{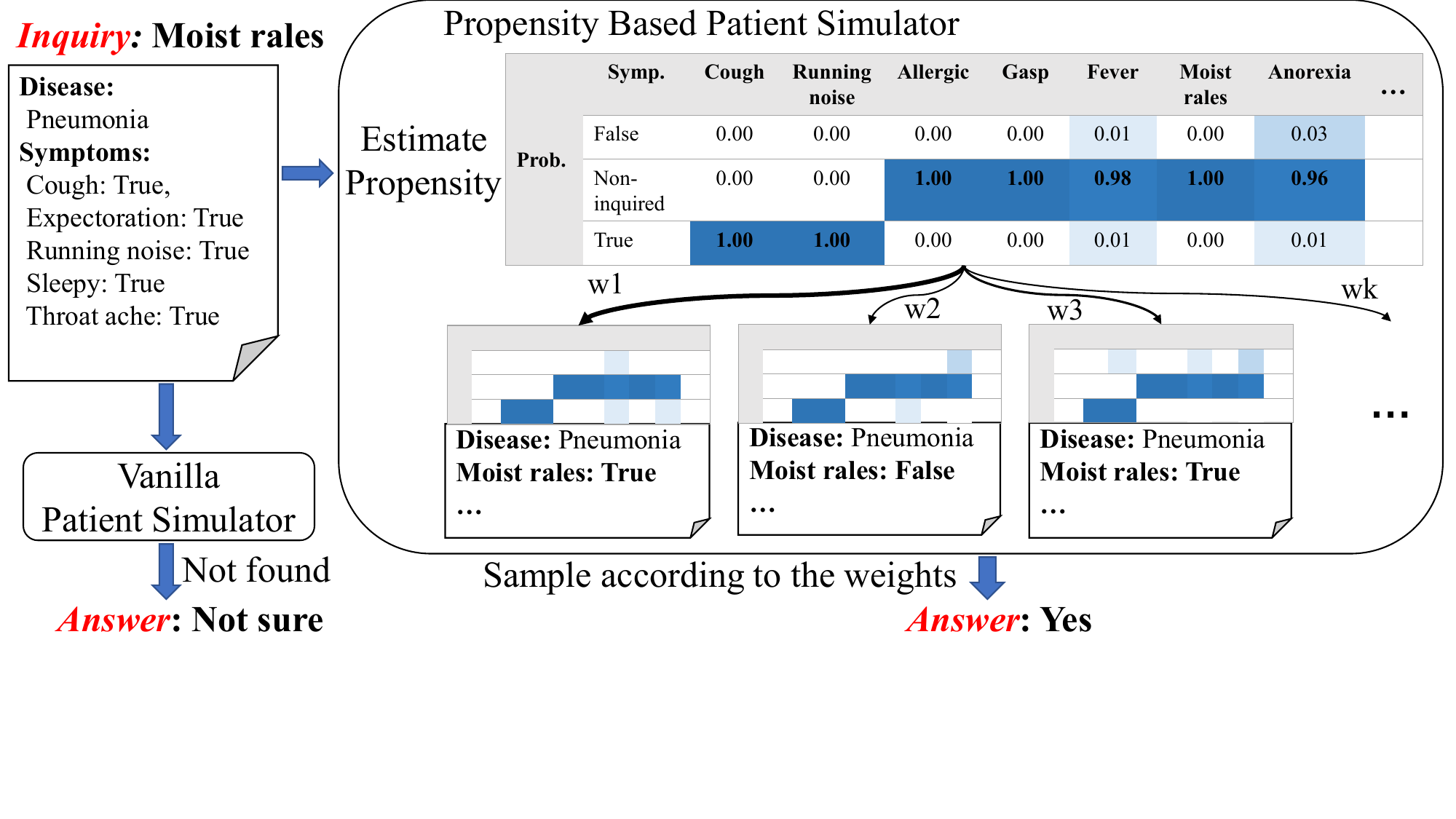}
	\caption{A running example of PBPS*. The vanilla patient simulator can only answer counterfactual symptom inquiries with `not sure' words. PBPS calculates the similarity weights of other records according to the propensity scores and then it samples an answer from one of the similar records.}
	\label{fig:example_pbps}
\end{figure*}

\begin{figure*}[!t]
	\centering
	\includegraphics[clip=true, trim= 0pt 205pt 180pt 0pt, width=0.7\linewidth]{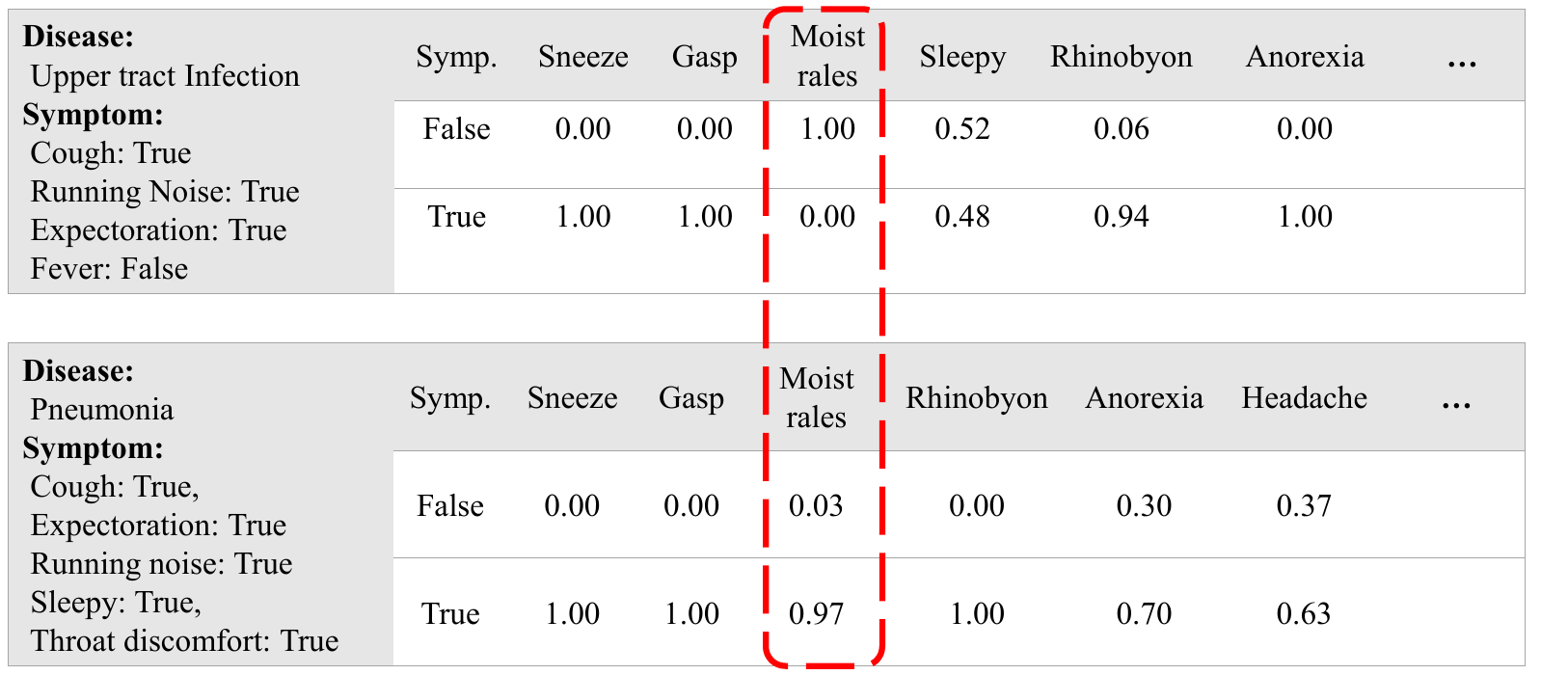}
	\caption{Answer probabilities of several counterfactual symptom inquiries of PBPS w.r.t. two different diagnosis records. Both of the records have respiratory diseases and therefore they have many symptoms in common. However, PBPS captures the critical and distinguishable symptom, i.e., `moist rales', between these diseases.}
	\label{fig:moist_example}
\end{figure*}

\begin{figure*}[t]
	\centering
	\includegraphics[clip=true, trim= 0pt 370pt 0pt 0pt, width=\linewidth]{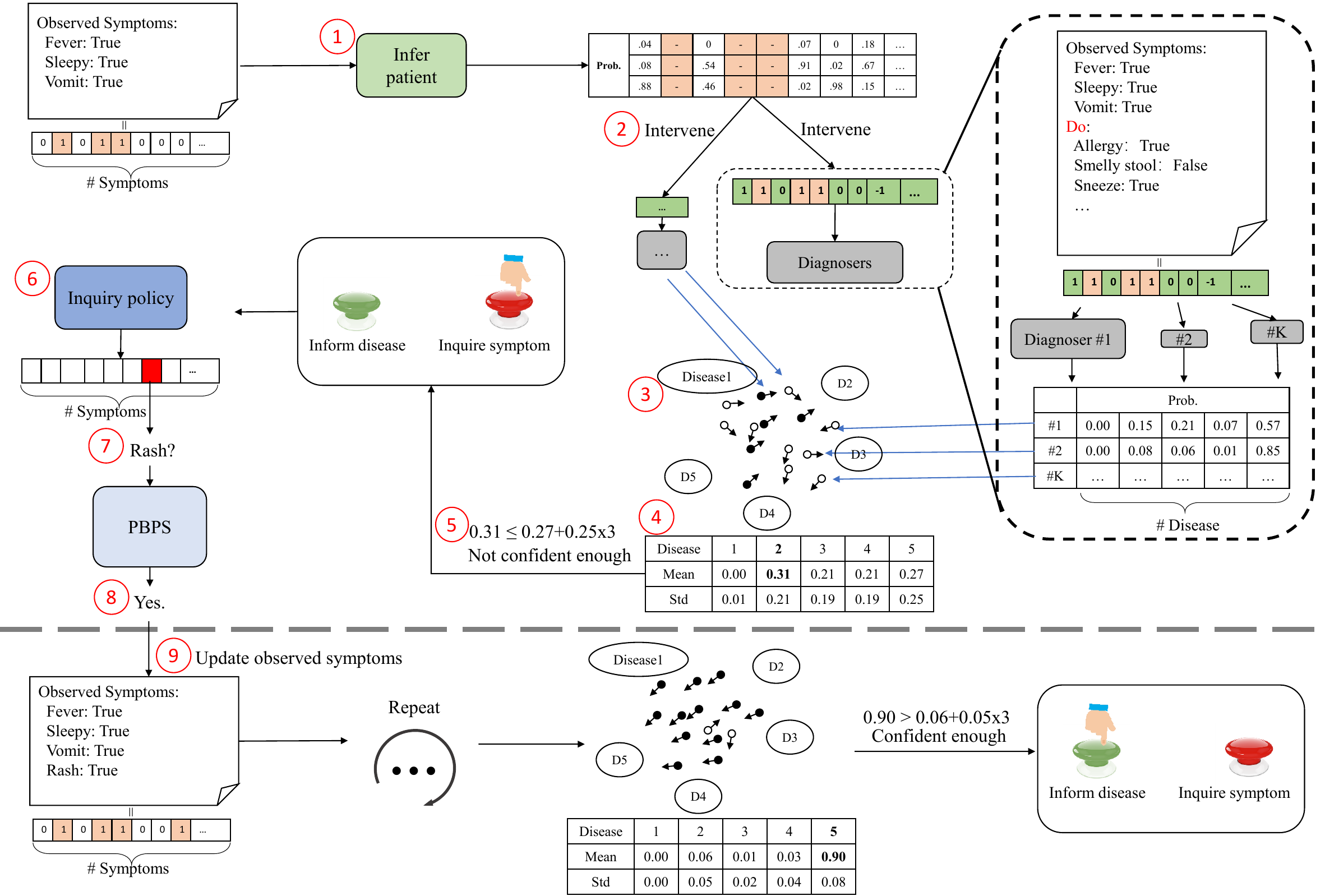}
	\vspace{-10pt}
	\caption{A running example of P2A. For better understanding, we mark the execution order with red circled numbers. The diagnosis branch takes $\mathbf{s}_t^v$ to draw $K$ final states which are fed into the $B$ bootstrapping diagnosers to obtain the expectation and deviation. P2A keeps inquiring about symptoms of the patient simulator until the diagnosis meets the decision threshold.}
	\label{fig:example_p2a}
\end{figure*}

\textbf{Performance improvement of MDA agents.} As the answers from PBPS are more realistic, we have verified whether PBPS can be more beneficial to existing RL baselines than the original simulator. In specific, we take $\mathcal{P}^{\rm train}$ to interact with the DQN and KR-DQN agents to train their policies, then, we evaluate their success rates on the diagnosis episodes generated by completing the absent value of the symptoms in the test set of the MDA benchmarks. 
Fig.~\ref{fig:usersim_cmp} showcases the comparisons between the PS and PBPS based on success rates across RL baselines and benchmarks. We have observed that when trained with the episodes generated by PBPS, the RL baselines reap a significant performance over the original. It evidences that using PBPS helps to develop more competitive MDA agents for real-world applications.

\textbf{Comparison with advanced RL methods}
Different from RL formulation, our P2A leverages the decision threshold to modulate between discrimination and inquiry. Although the evaluations on MDA tasks have witnessed great success, there are various domain-specific factors, e.g. lack of ground-truth patient simulator, that could cause the results difficult to analyze. To this end, one might be interested in a more direct performance of P2A in solving the sequential discriminant problem in comparison to advanced RL methods.

To see the difference between P2A and RL methods directly, we design a symbolic and difficult enough task, named ``mystery boxes", for sequential discrimination and compare our approach with different advanced RL methods. In this task, there will be $N$ opaque boxes with identical appearances. And inside each box, there is a unique note of a binary array with a length of 10 following its box ID $\in [1,..,N]$. And the agent is asked to guess the box ID after interactively inquiring about the information on the note of each box. For each inquiry, the agent can only access one bit of the binary note. To introduce randomness, at the beginning of the interaction with a box, there is a 30\% chance for each bit of the box note to flip its value (from 0 to 1 or from 1 to 0). The agent can choose to inquire about a bit of the note from the box or stop to inform the box ID. Each inquiry action will cost the agent a penalty $r_t=-0.1$. And the agent will receive a positive/negative reward $r_T=1/-1$ for correct/incorrect discrimination. In this task, we can easily increase the difficulty of the task by increasing $N$. One can see medical diagnosis as the special case of the mystery boxes.

In our new experiment, we use the open-release RL library Tianshou (https://github.com/thu-ml/tianshou) to conduct experiments with DQN~\cite{mnih2015human}, A2C~\cite{mnih2016asynchronous}, PPO~\cite{schulman2017proximal}, and our P2A/P2A-DT. All network architectures are 4-layered MLP. The number of training episodes for each method is $N \times 1e3$. And the learning rate is $1e-4$. For the other hyperparameters, we use the default values used in Tianshou and our P2A. In our experiments, we compare different methods in terms of N and R. During testing, each box appears 50 times. The averaged prediction accuracies of different methods are reported in the following.

\begin{table}[t]
\centering
\caption{The experiment results of different RL-based methods and our P2A on the task of mystery boxes with different $N$.}
\begin{tabular}{c|cccc} \hline
Methods & N=8 & N=16 & N=32 & N=64 \\ \hline
DQN~\cite{mnih2015human}     & 0.95\textpm 0.03	& 0.65\textpm 0.08	& 0.08\textpm 0.02 &	0.02\textpm 0.01 \\
A2C~\cite{mnih2016asynchronous}     &0.97\textpm 0.02	& 0.81\textpm 0.11	& 0.32\textpm 0.15 &	0.12\textpm 0.04 \\
PPO~\cite{schulman2017proximal}    & 0.97\textpm 0.02	& 0.83\textpm 0.10	& 0.36\textpm 0.11 &	0.12\textpm 0.03 \\ \hline
P2A     &  0.97\textpm 0.02	& 0.87\textpm 0.07	& 0.71\textpm 0.11 &	0.55\textpm 0.19 \\
P2A-DT  &  0.97\textpm 0.02	& 0.95\textpm 0.03	& 0.94\textpm 0.03 &	0.89\textpm 0.10 \\ \hline
\end{tabular}
\end{table}

When $N$ becomes larger, the model needs more information to distinguish different boxes, which leads to more inquiry punishment. We find that when $N$ is larger than 32, the RL method has a significant decline, while our method can still maintain high accuracy. Even when it reaches 64, only P2A/P2A-DT can still maintain a high classification accuracy, and P2A-DT can still maintain very high accuracy.

The potential reason why RL baselines drop severely as $N$ increases is that, when $N$ is large, the discriminations at the beginning of the training period are likely wrong. Therefore, the RL-based methods will be frequently punished at the beginning of the training period, which leads the policy to improve to reduce the punishment with fewer inquiries and rush discrimination. Differently, our P2A can still obtain good performance as $N$ increases. The reason is that the inquiry will search for an increase in the confidence to meet the decision threshold. Therefore, it can not jump to a conclusion to avoid short-sighted punishment. Especially for those cases where the decision threshold is met, the accuracy is considerably improved.

From this set of experiments, we can see that, in the context of sequential discrimination, P2A has advantages over pure RL-based methods in the perspectives of both the optimization target and the experimental performance in terms of different levels of task difficulty.

\begin{table}[t]
\centering
\caption{The ablation results of using random/causal mask during the training process under both In./Out. settings.}
\label{tab:mask}
\begin{tabular}{c|cc|cc}
\hline 
Data       & \multicolumn{2}{c|}{MZ} & \multicolumn{2}{c}{DX} \\ \hline
Setting    & \textbf{Out.}        & \textbf{In.}        & \textbf{Out.}        & \textbf{In.}        \\ \hline
P2A-random mask	& 0.77\textpm 0.03 &	0.87\textpm 0.03 &	0.92\textpm 0.03 &	0.92\textpm 0.02 \\
P2A	& \textbf{0.78\textpm 0.04} &	\textbf{0.89\textpm 0.03} &	\textbf{0.93\textpm 0.02} &	\textbf{0.94\textpm 0.02}  \\ \hline
\end{tabular}
\end{table}

\textbf{Mask selection.}
In our P2A, we use the visitation mask  $\mathbf{m}_t$ to create counterfactual training data for training the  propensity latent feature extractor of P2A. Since this strategy follows the Granger causality, we name our mask strategy the causal mask. In order to testify to the effect of the mask, we have designed a new series of experiments. Specifically, in the training process, the “P2A-random mask” uses the simulator trained with the random mask, while in the test stage, we still used the simulator trained with the causal mask. The results are in Tab.~\ref{tab:mask}. From the results, we can see that when the mask strategies used in our test and training are different, the performance will be affected. But we also noticed that the decline was not obvious. This implies that the order of inquiries plays a small part in determining the answers of the patients for different symptoms.

\section{Running Examples}
\textbf{Example of PBPS.} To better understand the workflow of PBPS, we have also added some numerical examples to illustrate visually as shown in Fig.~\ref{fig:example_pbps}. Since our PBPS uses latent features for matching, which is hard to visually understand, we instead use an equivalent substitute of PBPS, i.e., PBPS*, for demonstration. From the figure, we inspect what has happened inside the PS/PBPS* when the symptom ``moist rales” is inquired about. The previous patient simulator PS simply searches the symptom in the anchor record and returns ``not sure” when the symptom is not found. As for PBPS*, when the symptom is not found in the record, PBPS* matches other records with the same disease according to their propensity scores and assigns weights to matched records according to the similarities. And finally, the symptom existence is sampled according to the similarity weights. From the example, we can see that the propensity of `non-inquired’ is about 1.0 for the symptom `moist rales’. This also reflects that patient simulators like PS and GEN are likely to answer with ``not sure”.

When inspecting the probabilities of answers to different symptom inquiries by PBPS, we found an interesting case as bounded by the red box in the tables of Fig.~\ref{fig:moist_example}. We found that between the two cases with respiratory diseases, i.e., ``upper tract infection” and ``pneumonia”, the probabilities of ``moist rales” differ toward two opposite extremes when the others are closed. Moist rale is a symptom that happens in the lung which is positioned in the lower tract. Therefore, the upper tract infection certainly does not include moist rales. The doctor can easily exclude the upper tract infection if the patient has moist rales. However, there is still a small chance (0.03) that the patient with pneumonia would not have moist rales. For example, if the secretions are little and almost existed in the alveolar cavity of the pneumonia patient, the doctor usually could not hear the moist rales. Although the moist rale is distinctive, the previous simulated patient with the upper tract infection will answer with ``not sure” in 99.18\% of the time, because the moist rales in 99.18\% of the records with the upper tract infection are not non-inquired.

\begin{figure*}[t]
	\centering
	\includegraphics[clip=true, trim=0pt 10pt 0pt 0pt, width=\linewidth]{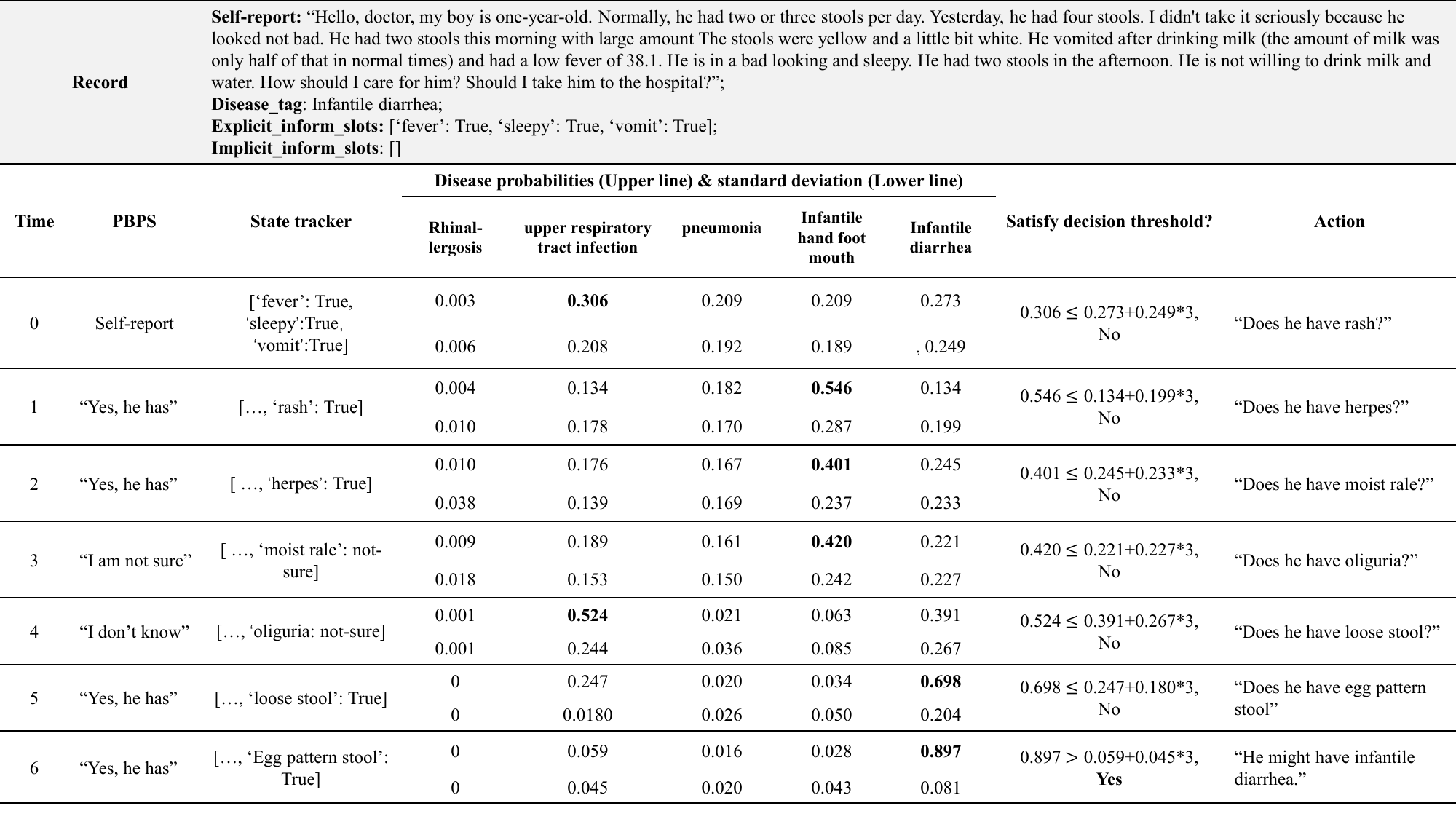}
	\caption{A correct diagnosis example of our methods on DX. As the diagnosis progresses, the expected disease is changing accordingly, and finally, the correct diagnosis is made with DT satisfied.}
	\label{fig:good_example}
\end{figure*}

\textbf{Example of P2A.} To intuitively demonstrate our P2A, Fig.~\ref{fig:example_p2a} is an intermediate running example. This figure shows the working pipeline of our framework from the beginning to the end. At each turn, P2A infers the possible complete symptoms from the observed, and then it conducts do-operation to the non-inquired symptoms. After that, the observed and the estimated symptoms are fed to several bootstrapping diagnosers to estimate the diagnostic probabilities and confidences. If the decision threshold is not met, P2A will continue inquiring. Otherwise, it stops to inform disease. This example shows that P2A is unconfident in the beginning and becomes very confident in the end. To keep the example in brief, we don’t draw the complete process here and more numerical results of this example have been summarized in Fig.~\ref{fig:good_example}.

\begin{figure*}[!t]
	\centering
	\includegraphics[clip=true, trim=0pt 60pt 0pt 0pt, width=\linewidth]{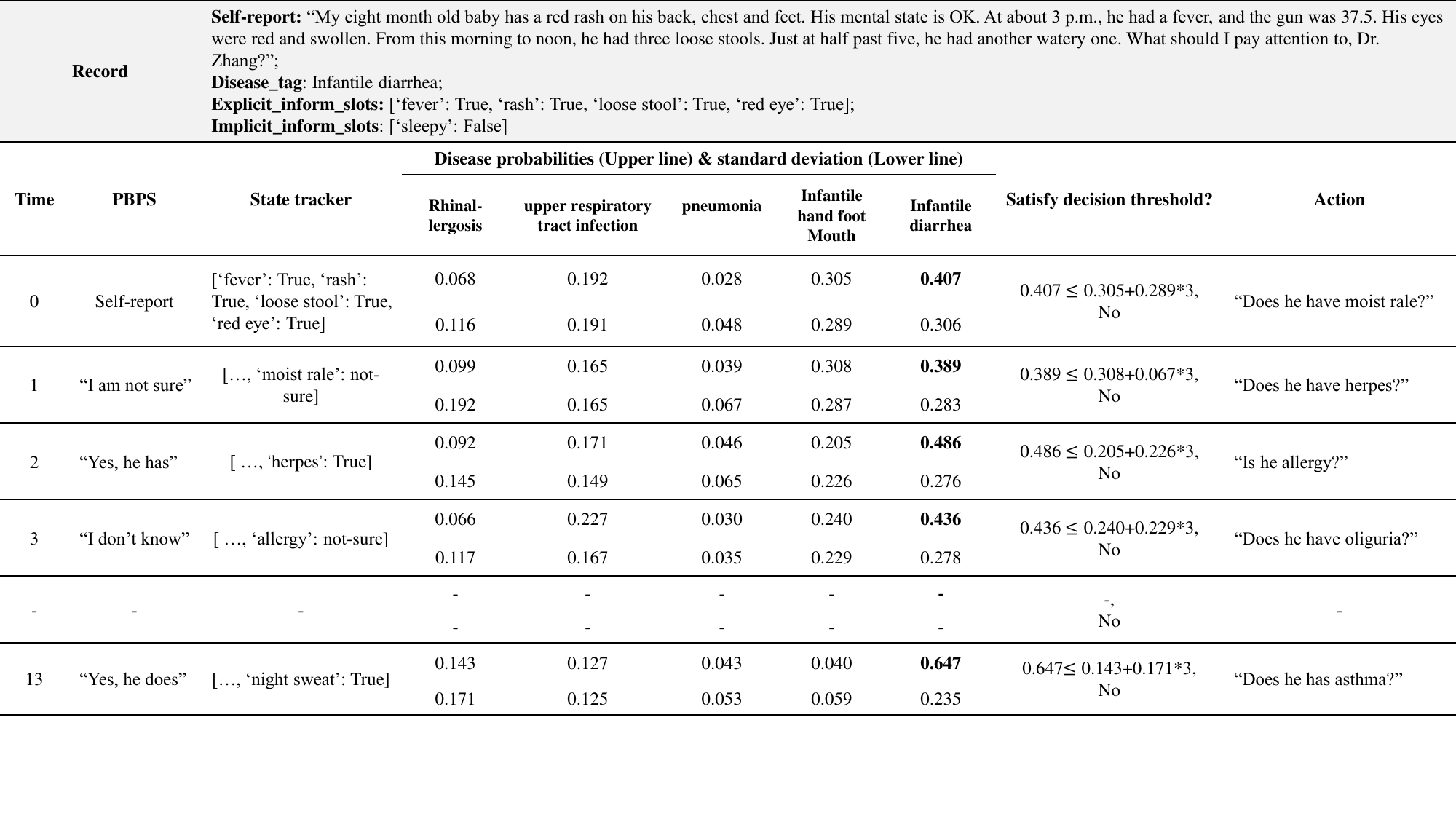}
	\caption{A diagnosis example without satisfying DT produced by our methods on DX. The disease is correctly estimated at the very beginning but the agent is forced to query more symptoms in order to meet DT. The most likely diseases of items (from Time 4 to Time 13) are still 'Infantile diarrhea', therefore, we omitted them to keep the demonstration brief.}
	\label{fig:bad_example}
\end{figure*}

Fig.~\ref{fig:good_example} consists of the anchor record and a series of interaction items. The interaction item at each time point includes the output of the patient simulator PBPS, the mentioned symptoms, the estimated disease expected probability and standard deviation, whether the decision threshold is met, and the actual action of the agent P2A. As shown in Fig.~\ref{fig:good_example}, in the beginning, the patient simulator reported its explicit symptoms to the agent, according to which the agent predicted that upper respiratory tract infection is the most likely disease. But the agent also foresaw the future variables and its decision threshold suggested that the disease could also be infantile diarrhea if some relevant symptoms were presented. As a result, our agent was required to keep inquiring, and during diagnosis, the most likely disease was also changing. Finally, when the decision threshold was reached, the agent made a confident and correct diagnosis. This example shows that simply rendering the most likely disease without considering future possibilities can lead to reckless and erroneous results. In addition, from the perspective of the patient simulator, we can also observe that the set of implicit symptoms is empty in the anchor record. Therefore, if the agent were interacting with the original patient simulator, it cannot obtain useful information.

However, our method also has side effects, as shown in Fig.~\ref{fig:bad_example}. We observe that though the agent first estimates correctly based on the explicit symptoms, the agent is forced to query too many symptoms for meeting the decision threshold. Consequently, the dialogue became over-length. These two examples can also be seen in the real-life decision-making process. Sometimes a person is called cautious and responsible when making a decision rationally, while sometimes the same person is called wandering and indecisive when thinking too much and missing opportunities. Although most of the time, people don't want to be indecisive. But as for medical applications, if an MDA agent is hesitant, it is always better not to make a decision and seek experts for help than to make a wrong decision recklessly.

\section{Discussions and Limitations} 
In this section, we first discuss the further connection between PBPS and P2A. After, we discuss the limitations of our approach.

\subsection{Discussions}

\textbf{{PBPS Helps Disentangling Inquiry and Diagnosis.}}  Since the answers from the original patient simulator would be informative only when the inquiries are factual, state $\mathbf{s}_t$ is prone to be insufficient to estimate the disease in the early diagnosis process. It leads the MDA agent to estimate the disease only after the symptom-inquiry process, therefore the actions of symptom inquiry and disease informing can be assigned to a single policy~\cite{wei2018task,xu2019end}. Conversely, with our PBPS, an agent is able to gather an informative response per step. Hence, it is encouraged to learn a diagnoser to make and adjust its disease estimation along the symptom-inquiry process in parallel.

\textbf{Reject the record $q$ whose $y_a^{(q)} = 0$ during matching.} For clarity and simplicity, we denote the symptom as $a$, the indicator of whether $a$ had been inquired as $m_a$ (e.g., $m_a=1$ means that $a$ had been inquired about), the state of the patient as $z$ ($z$ includes all recorded symptoms and the disease label), and the answer from the patient as $o_a$. In MDA, $m_a = \mathbb{I}(y_a^{(q)} \neq 0)$. Therefore, for all the inquired symptoms recorded in the dataset, their statuses $m_a$ are equal to 1. By this denotation, the question about whether to reject records $q$ whose $y_a^{(q)} = 0$ can be formulated to ``which answer model should be used, $P(O_a | z, m_a=1)$ or $P(O_a | z)$"? In our context, i.e., building an interactive patient simulator, $P(O_a|z,m_a=1)$ is the model we shall use. This is because this model takes the inquiry into account and stands for the patient's response after the symptom is inquired about.

\subsection{Limitations}

\textbf{Extending to new diseases.} Incremental learning is one of the hottest topics in the literature. We believe that by introducing the techniques in the field of incremental learning, our model should be able to extend to handle new diseases without training from scratch.

And to investigate the extensibility of our model, we have designed a new series of experiments for studying the incremental learning ability of our P2A. The experiment includes two stages, i.e., the subset diseases pretraining stage and the full diseases finetuning stage. In the first stage, we randomly select a disease and removed all records of the selected disease from the training set. Then, P2A is trained to diagnose the rest diseases. In the second stage, training records of the removed disease are added back to the training set, and the P2A is finetuned with the new data. Specifically, before finetuning, the last layer of each bootstrapping diagnoser and the first layer of the policy network are replaced with newly initialized layers to fit the new set of diseases. During finetuning, we fixed the old parameters and only updated the newly initialized layers. After converging, the old parameters are unfrozen and further updated with the new layers jointly.

In this set of experiments, we are interested in how many previous disease records are required so that P2A can finetune to address new diseases without forgetting how to diagnose the previous diseases. To this end, we adopt 3 settings in terms of the different numbers of anchor records of each previous disease, i.e., 5, 10, and full shots. We don’t take 0 shot into consideration because 0 shot means P2A only finetunes with the new disease data and there will be no need for discrimination. As for evaluation, at the subset diseases training stage, P2A is evaluated using the subset of test records without the selected disease, and at the full diseases finetuning stage, P2A is evaluated using all test records. We randomly select “infantile diarrhea” as the removed disease at the first stage for both DX and MZ without special reasons.
\begin{table}[t]
\centering
\caption{The incremental learning experiments of P2A which is finetuned to handle a new disease.}
\label{tab:il}
\setlength{\tabcolsep}{3pt}
\begin{tabular}{c|c|c|ccc}
\hline
\multirow{2}{*}{Dataset} & \multirow{2}{*}{Disease} & \multirow{2}{*}{Pretrain stage} & \multicolumn{3}{c}{Finetune stage} \\ \cline{4-6} 
                         &                          &                           & 5      & 10      & full      \\ \hline
\multirow{3}{*}{DX}      & subset                   &   0.96\textpm 0.01                       &  0.83\textpm 0.12       &    0.91\textpm 0.07    &     0.93\textpm 0.01       \\
                         & new disease              &     -                      &  0.97\textpm 0.03       &   0.96\textpm 0.02       &     0.91\textpm 0.02       \\
                         & Average                  &     0.96\textpm 0.01                       &  0.86\textpm 0.11       &    0.92\textpm 0.06      &    0.93\textpm 0.01        \\ \hline
\multirow{3}{*}{MZ}      & subset                   &      0.92\textpm 0.01                      &    0.73\textpm 0.17    &   0.83\textpm 0.06      &       0.90\textpm 0.01     \\
                         & new disease              &      -                     &     0.95\textpm 0.04   &     0.92\textpm 0.03    &   0.87\textpm 0.02        \\
                         & Average                  &     0.92\textpm 0.01                       &   0.78\textpm 0.15      &   0.85\textpm 0.05       &   0.89\textpm 0.01       \\ \hline
\end{tabular}
\end{table}

From the results in Tab.~\ref{tab:il}, we can observe that, after introducing a new disease, P2A can still obtain good performance in diagnosing previous methods with about 5 training records of each previous disease. And the performance of diagnosing previous diseases increases as more training records are used. As for the new disease, we observe that when there are 5 records of previous diseases, P2A can learn to focus on diagnosing the new disease while obtaining less balanced performance on other diseases. When more records of other diseases are added, the performance of P2A becomes more balanced across different diseases. From this experiment, we can see that P2A does not require re-train all data from scratch when new diseases are added. However, the performance might be imbalanced if only a few previous records are used when training with the new diseases. Nevertheless, we believe that with more advanced incremental learning techniques and more advanced network architecture, the performance can be further improved.

\textbf{Enumerable inquiries and answers}. In task MDA, there are three types of answers, i.e. "Yes", "No" and "Not sure". And the candidate inquiries are equaled to the types of symptoms. Knowing these candidate answers allows our intervener to intervene in future scenarios. However, in real-life cases, the dialogue could be more complex. Especially, the inquiries and the answers are open-vocabulary. To resolve this issue in more complex tasks, one might resort to modeling the action distribution and samples candidate actions from the distribution.

\textbf{Conservatism}. Our decision threshold imposes strict regulation on the inquiry branch. However, sometimes it could happen that the decision threshold isn't satisfied for a long time though the intermediate diagnosis result is correct. Consequently, the dialogue became over-length. A negative example is illustrated in Appendix D. Nevertheless, for a medical application, it should be always better for the diagnostic agent to make a safer and more comprehensive decision at the cost of more interactions, rather than making a wrong and reckless decision. And human experts could always intervene when the diagnostic agent is hesitant to make a decision. In practice, we force the agent to stop inquiring if the length of discoursing approaches the maximum step.



\bibliographystyle{IEEEtran}

\begin{thebibliography}{10}
\providecommand{\url}[1]{#1}
\csname url@samestyle\endcsname
\providecommand{\newblock}{\relax}
\providecommand{\bibinfo}[2]{#2}
\providecommand{\BIBentrySTDinterwordspacing}{\spaceskip=0pt\relax}
\providecommand{\BIBentryALTinterwordstretchfactor}{4}
\providecommand{\BIBentryALTinterwordspacing}{\spaceskip=\fontdimen2\font plus
\BIBentryALTinterwordstretchfactor\fontdimen3\font minus
  \fontdimen4\font\relax}
\providecommand{\BIBforeignlanguage}[2]{{%
\expandafter\ifx\csname l@#1\endcsname\relax
\typeout{** WARNING: IEEEtran.bst: No hyphenation pattern has been}%
\typeout{** loaded for the language `#1'. Using the pattern for}%
\typeout{** the default language instead.}%
\else
\language=\csname l@#1\endcsname
\fi
#2}}
\providecommand{\BIBdecl}{\relax}
\BIBdecl

\bibitem{mnih2015human}
V.~Mnih, K.~Kavukcuoglu, D.~Silver, A.~A. Rusu, J.~Veness, M.~G. Bellemare,
  A.~Graves, M.~Riedmiller, A.~K. Fidjeland, G.~Ostrovski \emph{et~al.},
  ``Human-level control through deep reinforcement learning,'' \emph{Nature},
  vol. 518, no. 7540, p. 529, 2015.

\bibitem{sutton1999policy}
S.~R. Sutton, A.~D. Mcallester, P.~S. Singh, and Y.~Mansour, ``Policy gradient
  methods for reinforcement learning with function approximation,'' in
  \emph{NeurIPS}, 1999, pp. 1057--1063.

\bibitem{xu2019end}
L.~Xu, Q.~Zhou, K.~Gong, X.~Liang, J.~Tang, and L.~Lin, ``End-to-end
  knowledge-routed relational dialogue system for automatic diagnosis,'' in
  \emph{AAAI}, 2019.

\bibitem{wei2018task}
Z.~Wei, Q.~Liu, B.~Peng, H.~Tou, T.~Chen, X.~Huang, K.-F. Wong, and X.~Dai,
  ``Task-oriented dialogue system for automatic diagnosis,'' in \emph{ACL},
  vol.~2, 2018, pp. 201--207.

\bibitem{cole2010illustrating}
S.~R. Cole, R.~W. Platt, E.~F. Schisterman, H.~Chu, D.~Westreich,
  D.~Richardson, and C.~Poole, ``Illustrating bias due to conditioning on a
  collider,'' \emph{IJE}, vol.~39, no.~2, pp. 417--420, 2010.

\bibitem{rubin1974estimating}
D.~B. Rubin, ``Estimating causal effects of treatments in randomized and
  nonrandomized studies.'' \emph{JEP}, vol.~66, no.~5, p. 688, 1974.

\bibitem{neyman1923applications}
J.~Neyman, ``Sur les applications de la th{\'e}orie des probabilit{\'e}s aux
  experiences agricoles: Essai des principes,'' \emph{Roczniki Nauk
  Rolniczych}, vol.~10, pp. 1--51, 1923.

\bibitem{dehejia2002propensity}
R.~H. Dehejia and S.~Wahba, ``Propensity score-matching methods for
  nonexperimental causal studies,'' in \emph{RES}, vol.~84.\hskip 1em plus
  0.5em minus 0.4em\relax MIT Press, 2002.

\bibitem{austin2011introduction}
P.~C. Austin, ``An introduction to propensity score methods for reducing the
  effects of confounding in observational studies,'' \emph{MBR}, vol.~46,
  no.~3, pp. 399--424, 2011.

\bibitem{rosenbaum1983central}
P.~R. Rosenbaum and D.~B. Rubin, ``The central role of the propensity score in
  observational studies for causal effects,'' in \emph{Biometrika}.\hskip 1em
  plus 0.5em minus 0.4em\relax Oxford University Press, 1983.

\bibitem{Pearl2012TheDr}
J.~Pearl, ``The do-calculus revisited,'' in \emph{UAI}, 2012.

\bibitem{ratcliff2016diffusion}
R.~Ratcliff, P.~L. Smith, S.~D. Brown, and G.~McKoon, ``Diffusion decision
  model: Current issues and history,'' \emph{Trends in cognitive sciences},
  vol.~20, no.~4, pp. 260--281, 2016.

\bibitem{pearl2009causal}
J.~Pearl \emph{et~al.}, ``Causal inference in statistics: An overview,''
  \emph{Statistics surveys}, vol.~3, pp. 96--146, 2009.

\bibitem{bollen2013eight}
K.~A. Bollen and J.~Pearl, ``Eight myths about causality and structural
  equation models,'' in \emph{Handbook of causal analysis for social
  research}.\hskip 1em plus 0.5em minus 0.4em\relax Springer, 2013, pp.
  301--328.

\bibitem{chen2019counterfactual}
L.~Chen, H.~Zhang, J.~Xiao, X.~He, S.~Pu, and S.-F. Chang, ``Counterfactual
  critic multi-agent training for scene graph generation,'' in \emph{CVPR},
  2019, pp. 4613--4623.

\bibitem{kaihua2020unbiased}
T.~Kaihua, N.~Yulei, H.~Jianqiang, S.~Jiaxin, and Z.~Hanwang, ``Unbiased scene
  graph generation from biased training,'' in \emph{CVPR}, 2020.

\bibitem{qi2020two}
J.~Qi, Y.~Niu, J.~Huang, and H.~Zhang, ``Two causal principles for improving
  visual dialog,'' in \emph{CVPR}, 2020, pp. 10\,860--10\,869.

\bibitem{wang2020visual}
T.~Wang, J.~Huang, H.~Zhang, and Q.~Sun, ``Visual commonsense r-cnn,'' in
  \emph{CVPR}, 2020, pp. 10\,760--10\,770.

\bibitem{abbasnejad2020counterfactual}
E.~Abbasnejad, D.~Teney, A.~Parvaneh, J.~Shi, and A.~v.~d. Hengel,
  ``Counterfactual vision and language learning,'' in \emph{CVPR}, 2020.

\bibitem{yu2021reinforcement}
C.~Yu, J.~Liu, S.~Nemati, and G.~Yin, ``Reinforcement learning in healthcare: A
  survey,'' \emph{ACM Computing Surveys (CSUR)}, vol.~55, no.~1, pp. 1--36,
  2021.

\bibitem{dasgupta2018causal}
I.~Dasgupta, X.~J. Wang, S.~Chiappa, J.~Mitrovic, A.~P. Ortega, D.~Raposo,
  E.~Hughes, P.~Battaglia, M.~Botvinick, and Z.~Kurth-Nelson, ``Causal
  reasoning from meta-reinforcement learning,'' in \emph{ICLR}, 2019.

\bibitem{oberst2019counterfactual}
M.~Oberst and D.~Sontag, ``Counterfactual off-policy evaluation with gumbel-max
  structural causal models,'' in \emph{ICML}, 2019.

\bibitem{foerster2018counterfactual}
N.~J. Foerster, G.~Farquhar, T.~Afouras, N.~Nardelli, and S.~Whiteson,
  ``Counterfactual multi-agent policy gradients,'' in \emph{NCAI}, 2018.

\bibitem{bareinboim2016causal}
E.~Bareinboim and J.~Pearl, ``Causal inference and the data-fusion problem,''
  in \emph{Proceedings of the National Academy of Sciences}, vol. 113.\hskip
  1em plus 0.5em minus 0.4em\relax National Academy Sciences, 2016, pp.
  7345--7352.

\bibitem{schatzmann2007agenda}
J.~Schatzmann, B.~Thomson, K.~Weilhammer, H.~Ye, and S.~Young, ``Agenda-based
  user simulation for bootstrapping a pomdp dialogue system,'' in
  \emph{ACL}.\hskip 1em plus 0.5em minus 0.4em\relax ACL, 2007, pp. 149--152.

\bibitem{li2016user}
X.~Li, Z.~Lipton, B.~Dhingra, L.~Li, J.~Gao, and Y.-N. Chen, ``A user simulator
  for task-completion dialogues,'' in \emph{arXiv:1612.05688}, 12 2016.

\bibitem{pearl2009causality}
J.~Pearl, \emph{Causality}.\hskip 1em plus 0.5em minus 0.4em\relax Cambridge
  university press, 2009.

\bibitem{peng2018deep}
B.~Peng, X.~Li, J.~Gao, J.~Liu, and K.-F. Wong, ``Integrating planning for
  task-completion dialogue policy learning,'' in \emph{ACL}, 2018.

\bibitem{chen2017survey}
H.~Chen, X.~Liu, D.~Yin, and J.~Tang, ``A survey on dialogue systems: Recent
  advances and new frontiers,'' in \emph{SIGKDD}.\hskip 1em plus 0.5em minus
  0.4em\relax ACM, 2017.

\bibitem{stuart2010matching}
E.~A. Stuart, ``Matching methods for causal inference: A review and a look
  forward,'' \emph{SS}, vol.~25, no.~1, p.~1, 2010.

\bibitem{jakubowski2015latent}
M.~Jakubowski, ``Latent variables and propensity score matching: a simulation
  study with application to data from the programme for international student
  assessment in poland,'' \emph{Empirical Economics}, vol.~48, no.~3, pp.
  1287--1325, 2015.

\bibitem{he2022masked}
K.~He, X.~Chen, S.~Xie, Y.~Li, P.~Doll{\'a}r, and R.~Girshick, ``Masked
  autoencoders are scalable vision learners,'' in \emph{Proceedings of the
  IEEE/CVF Conference on Computer Vision and Pattern Recognition}, 2022, pp.
  16\,000--16\,009.

\bibitem{lipton2017bbq}
Z.~Lipton, X.~Li, J.~Gao, L.~Li, F.~Ahmed, and L.~Deng, ``Bbq-networks:
  Efficient exploration in deep reinforcement learning for task-oriented
  dialogue systems,'' in \emph{AAAI}, 2018.

\bibitem{li2017end}
X.~Li, Y.-N. Chen, L.~Li, J.~Gao, and A.~Celikyilmaz, ``End-to-end
  task-completion neural dialogue systems,'' in \emph{IJCNLP}, vol.~1, 2017.

\bibitem{madotto2018mem2seq}
A.~Madotto, C.-S. Wu, and P.~Fung, ``Mem2seq: Effectively incorporating
  knowledge bases into end-to-end task-oriented dialog systems.'' in
  \emph{ACL}, 2018, pp. 1468--1478.

\bibitem{wu2019global}
C.-S. Wu, R.~Socher, and C.~Xiong, ``Global-to-local memory pointer networks
  for task-oriented dialogue,'' in \emph{ICLR}, 2019.

\bibitem{lei2018sequicity}
W.~Lei, X.~Jin, M.-Y. Kan, Z.~Ren, X.~He, and D.~Yin, ``Sequicity: Simplifying
  task-oriented dialogue systems with single sequence-to-sequence
  architectures,'' in \emph{ACL}, vol.~1, 2018, pp. 1437--1447.

\bibitem{tang2016inquire}
K.-F. Tang, H.-C. Kao, C.-N. Chou, and E.~Y. Chang, ``Inquire and diagnose:
  Neural symptom checking ensemble using deep reinforcement learning,'' in
  \emph{NeurIPS}, 2016.

\bibitem{kao2018context}
H.-C. Kao, K.-F. Tang, and E.~Y. Chang, ``Context-aware symptom checking for
  disease diagnosis using hierarchical reinforcement learning,'' in \emph{AI},
  2018.

\bibitem{peng2018refuel}
Y.-S. Peng, K.-F. Tang, H.-T. Lin, and E.~Chang, ``Refuel: Exploring sparse
  features in deep reinforcement learning for fast disease diagnosis,'' in
  \emph{NeurIPS}, 2018, pp. 7333--7342.

\bibitem{10.1162/neco.2008.12-06-420}
R.~Ratcliff and G.~McKoon, ``{The Diffusion Decision Model: Theory and Data for
  Two-Choice Decision Tasks},'' \emph{Neural Computation}, vol.~20, no.~4, pp.
  873--922, 04 2008.

\bibitem{xia2020generative}
Y.~Xia, J.~Zhou, Z.~Shi, C.~Lu, and H.~Huang, ``Generative adversarial
  regularized mutual information policy gradient framework for automatic
  diagnosis,'' in \emph{AAAI}, 2020, pp. 1062--1069.

\bibitem{vanderweele2013three}
T.~J. VanderWeele, ``A three-way decomposition of a total effect into direct,
  indirect, and interactive effects,'' \emph{Epidemiology (Cambridge, Mass.)},
  vol.~24, no.~2, p. 224, 2013.

\bibitem{pearl1995causal}
J.~Pearl, ``Causal diagrams for empirical research. biometrika,'' in
  \emph{MathSciNet}, vol. 860, 1995.

\bibitem{koller2009probabilistic}
D.~Koller and N.~Friedman, \emph{Probabilistic graphical models: principles and
  techniques}.\hskip 1em plus 0.5em minus 0.4em\relax MIT press, 2009.

\bibitem{huang2012pearl}
Y.~Huang and M.~Valtorta, ``Pearl's calculus of intervention is complete,'' in
  \emph{CUAI}, 2012.

\bibitem{shpitser2012identification}
I.~Shpitser and J.~Pearl, ``Identification of conditional interventional
  distributions,'' in \emph{CUAI}, 2012.

\bibitem{mcallister2018robustness}
R.~McAllister, G.~Kahn, J.~Clune, and S.~Levine, ``Robustness to
  out-of-distribution inputs via task-aware generative uncertainty,'' in
  \emph{ICRA}, 2019.

\bibitem{pearl2014external}
J.~Pearl and E.~Bareinboim, ``External validity: From do-calculus to
  transportability across populations,'' \emph{Statistical Science}, pp.
  579--595, 2014.

\bibitem{whaite1997autonomous}
P.~Whaite and F.~P. Ferrie, ``Autonomous exploration: Driven by uncertainty,''
  \emph{TPAMI}, vol.~19, no.~3, pp. 193--205, 1997.

\bibitem{whaite1991uncertainty}
P.~{Whaite} and F.~P. {Ferrie}, ``From uncertainty to visual exploration,''
  \emph{TPAMI}, vol.~13, no.~10, pp. 1038--1049, 1991.

\bibitem{osband2018randomized}
I.~Osband, J.~Aslanides, and A.~Cassirer, ``Randomized prior functions for deep
  reinforcement learning,'' in \emph{NeurIPS}, 2018.

\bibitem{burda2018exploration}
Y.~Burda, H.~Edwards, A.~Storkey, and O.~Klimov, ``Exploration by random
  network distillation,'' in \emph{ICLR}, 2019.

\bibitem{brunke2022safe}
L.~Brunke, M.~Greeff, A.~W. Hall, Z.~Yuan, S.~Zhou, J.~Panerati, and A.~P.
  Schoellig, ``Safe learning in robotics: From learning-based control to safe
  reinforcement learning,'' \emph{Annual Review of Control, Robotics, and
  Autonomous Systems}, vol.~5, pp. 411--444, 2022.

\bibitem{pande2001six}
P.~S. Pande and L.~Holpp, \emph{What is six sigma?}\hskip 1em plus 0.5em minus
  0.4em\relax McGraw-Hill Professional, 2001.

\bibitem{parzen1960modern}
E.~Parzen, \emph{Modern probability theory and its applications}.\hskip 1em
  plus 0.5em minus 0.4em\relax John Wiley \& Sons, Incorporated, 1960.

\bibitem{wang2015dueling}
Z.~Wang, T.~Schaul, M.~Hessel, H.~Van~Hasselt, M.~Lanctot, and N.~De~Freitas,
  ``Dueling network architectures for deep reinforcement learning,'' in
  \emph{ICML}, 2016.

\bibitem{mnih2013playing}
V.~Mnih, K.~Kavukcuoglu, D.~Silver, A.~Graves, I.~Antonoglou, D.~Wierstra, and
  M.~Riedmiller, ``Playing atari with deep reinforcement learning,'' in
  \emph{CS}, 2013.

\bibitem{devlin2018bert}
J.~Devlin, M.-W. Chang, K.~Lee, and K.~Toutanova, ``Bert: Pre-training of deep
  bidirectional transformers for language understanding,'' \emph{arXiv preprint
  arXiv:1810.04805}, 2018.

\end{thebibliography}

\vfill

\vskip -2\baselineskip plus -1fil
\begin{IEEEbiography}[{\includegraphics[width=1in,height=1.25in,clip,keepaspectratio]{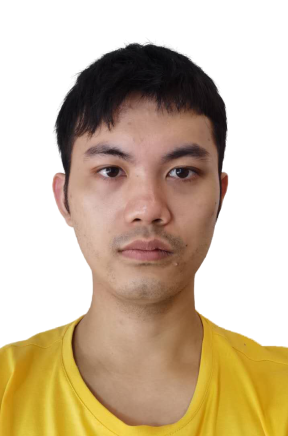}}]{Junfan Lin}
	received a B.S. degree in software engineering from Sun Yat-sen University, Guangzhou, China, where he is currently working toward a Ph.D. degree in computer
	science and technology, advised by Prof. L. Lin. His research interests include reinforcement learning and natural language processing.
\end{IEEEbiography}
\vskip -2\baselineskip plus -1fil
\begin{IEEEbiography}[{\includegraphics[width=1in,height=1.25in,clip,keepaspectratio]{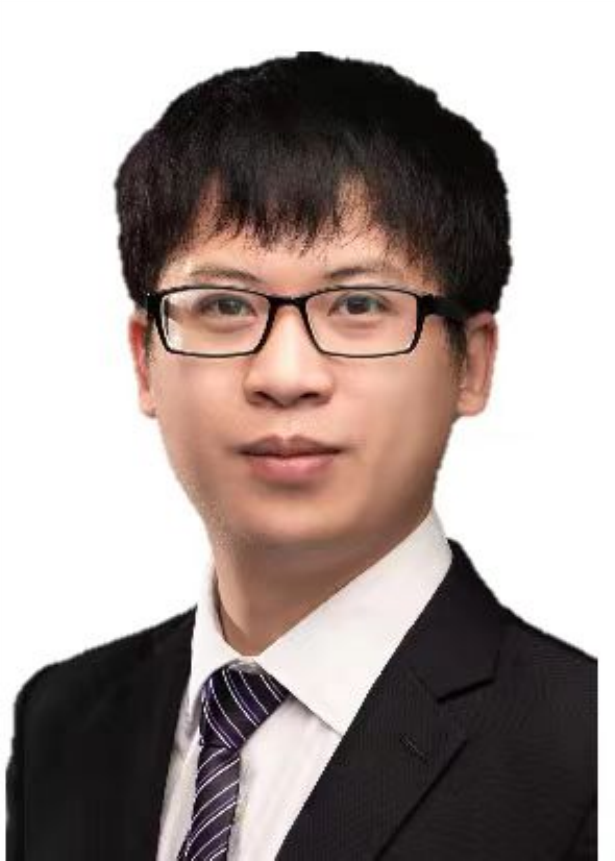}}]{Keze Wang} is nationally recognized as the Distinguished Young Scholars, currently serving as an Associate Professor at the School of Computer Science, Sun Yat-sen University, and a doctoral supervisor. He holds two Ph.D. degrees from Sun Yat-sen University (2017) and the Hong Kong Polytechnic University (2019). In 2018, he worked as a postdoctoral researcher at the University of California, Los Angeles, and returned to Sun Yat-sen University in 2021 as part of the "Hundred Talents Program." Dr. Wang has focused on reducing deep learning's dependence on training samples and mining valuable information from massive unlabeled data.
\end{IEEEbiography}
\vskip -2\baselineskip plus -1fil
\begin{IEEEbiography}[{\includegraphics[width=1in,height=1.25in,clip,keepaspectratio]{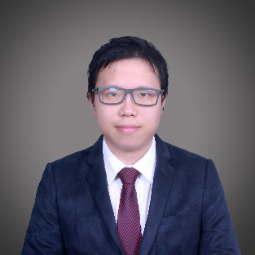}}]{Ziliang Chen}
	received a B.S. degree in mathematics and applied mathematics from Sun Yat-sen University, Guangzhou, China, where he obtained a Ph.D. degree in computer science and technology, advised by Prof. L. Lin. After that, he worked as an assistant professor in Jinan University. His research interests include computer vision and machine learning.
\end{IEEEbiography}
\vskip -2\baselineskip plus -1fil
\begin{IEEEbiography}[{\includegraphics[width=1in,height=1.25in,clip,keepaspectratio]{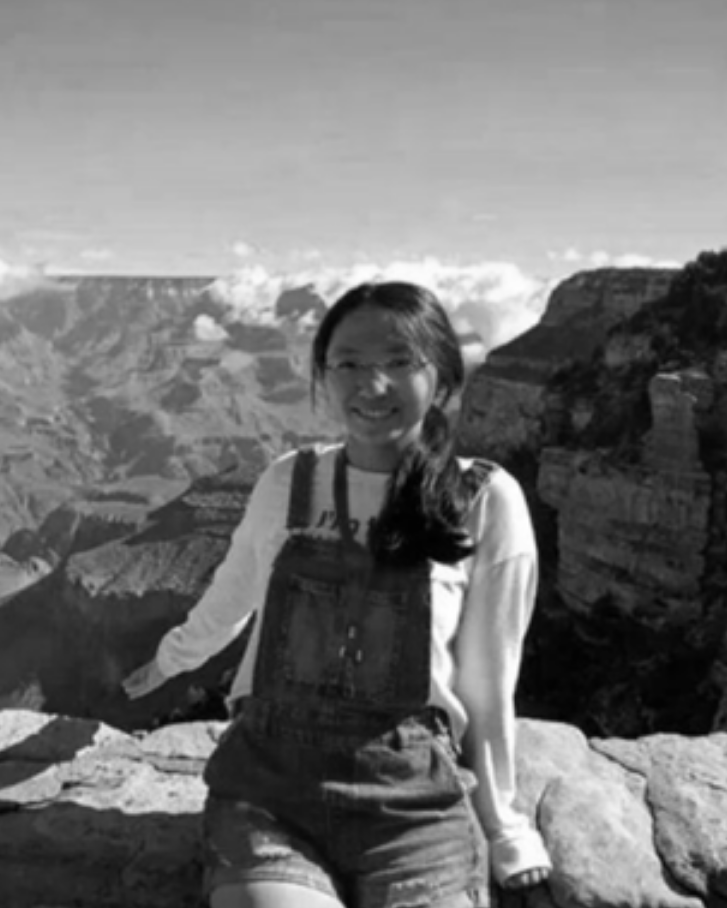}}]{Xiaodan Liang}
	received the PhD degree from Sun Yat-sen University, in 2016, advised by Liang Lin. She is an Associate Professor at Sun Yat-sen University. She served as an area chair of CVPR 2020, ICCV 2019 and WACV 2020. She was a postdoc researcher with Machine Learning Department, Carnegie Mellon University, working with Prof. Eric Xing, from 2016 to 2018. She has published several cutting-edge projects on human-related analysis including human parsing, pedestrian detection and instance segmentation, human pose estimation, and activity recognition.
\end{IEEEbiography}
\vskip -2\baselineskip plus -1fil
\begin{IEEEbiography}[{\includegraphics[width=1in,height=1.25in,clip,keepaspectratio]{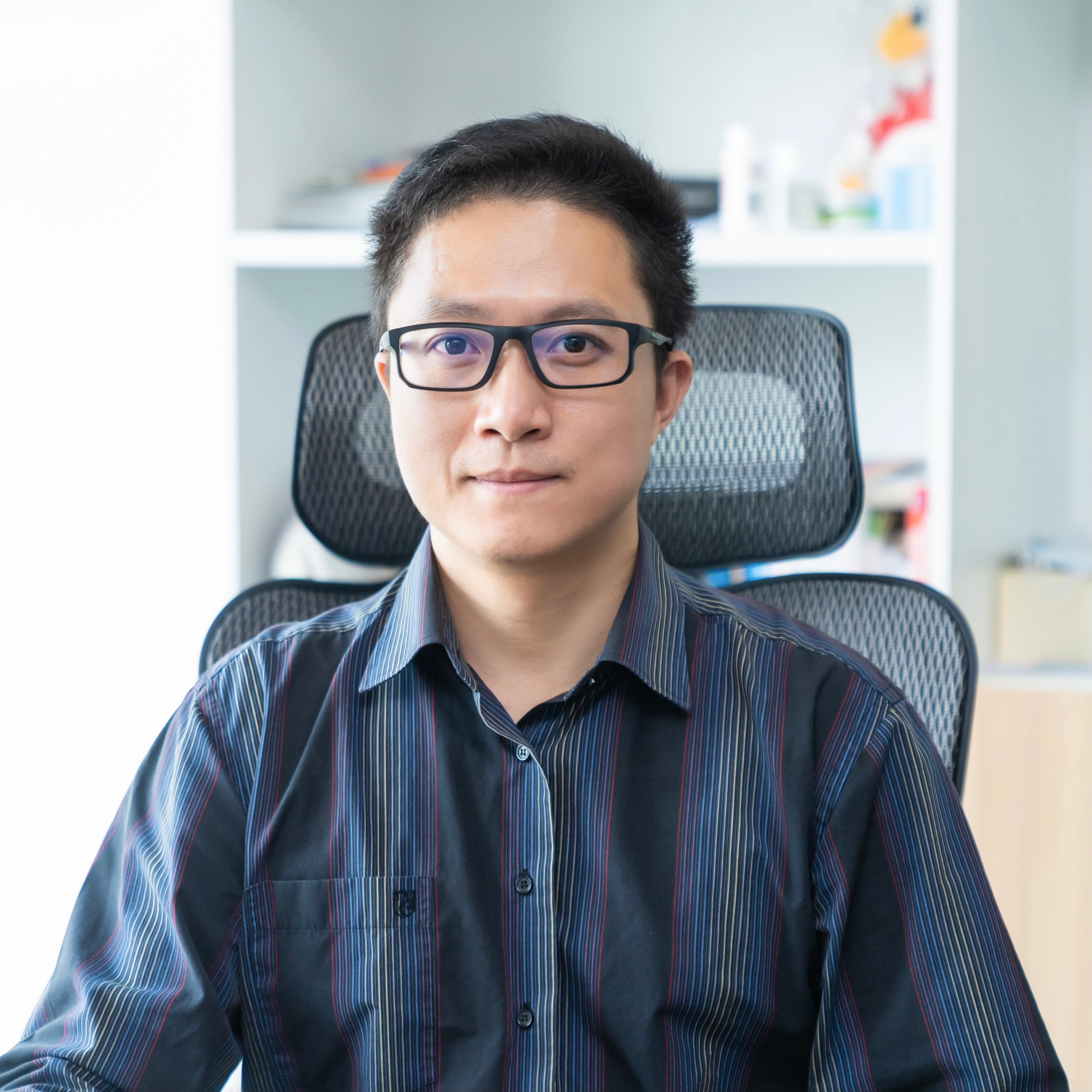}}]{Liang Lin} is a Full Professor of Sun Yat-sen University. He served as the Executive R\&D Director and Distinguished Scientist of SenseTime Group from 2016 to 2018, taking charge of transferring cutting-edge technology into products. He has authored or co-authored more than 200 papers in leading academic journals and conferences. He is an associate editor of IEEE Trans, Human-Machine Systems, and IET Computer Vision. He served as Area Chairs for numerous conferences such as CVPR, ICCV, and IJCAI. He is the recipient of numerous awards and honors including Wu Wen-Jun Artificial Intelligence Award, ICCV Best Paper Nomination in 2019, Annual Best Paper Award by Pattern Recognition (Elsevier) in 2018, Best Paper Dimond Award in IEEE ICME 2017, and Google Faculty Award in 2012. He is a Fellow of IET.
\end{IEEEbiography}




\vfill


\end{document}